\documentclass{article}

\usepackage{makecell}
\usepackage{tocloft} 
\usepackage{algorithm}
\usepackage{longtable}
\usepackage{algpseudocode}
\usepackage{PRIMEarxiv}
\usepackage{multirow}
\usepackage{amsmath}
\usepackage[utf8]{inputenc} % allow utf-8 input
\usepackage[T1]{fontenc}    % use 8-bit T1 fonts
\usepackage{hyperref}       % hyperlinks
\usepackage{url}            % simple URL typesetting
\usepackage{booktabs}       % professional-quality tables
\usepackage{amsfonts}       % blackboard math symbols
\usepackage{nicefrac}       % compact symbols for 1/2, etc.
\usepackage{microtype}      % microtypography
\usepackage{lipsum}
\usepackage{fancyhdr}       % header
\usepackage{graphicx}       % graphics
\usepackage{mathrsfs}
\graphicspath{{media/}}     % organize your images and other figures under media/ folder
\usepackage[authoryear]{natbib}
\usepackage[affil-it]{authblk} % For handling author affiliations
\usepackage{hyperref} % To enable clickable email links

\title{Enabling AI Scientists to Recognize Innovation: A Domain-Agnostic Algorithm for Assessing Novelty
}

\author[1,2]{Yao WANG}
\author[3]{Mingxuan CUI\textsuperscript{$\dagger$}}
\author[2]{Arthur JIANG}
\author[4]{Jun YAN}

\affil[1]{Department of Automation, Tsinghua University}
\affil[2]{Yidu Technology}
\affil[3]{Nankai University}
\affil[4]{Department of Biostatiatics, City University of Hong Kong}

\affil[ ]{\normalsize \href{mailto:wang-yao24@mails.tsinghua.edu.cn}{wang-yao24@mails.tsinghua.edu.cn}, 
\href{mailto:mingxuan.cui@mail.nankai.edu.cn}{mingxuan.cui@mail.nankai.edu.cn}, 
\href{mailto:ArthurSJiang@gmail.com}{ArthurSJiang@gmail.com}
\href{mailto:yan.jun@cityu.edu.hk}{yan.jun@cityu.edu.hk}}

\begin{document}
\maketitle

\renewcommand{\thefootnote}{\dag}
\footnotetext{Mingxuan CUI did this work during his internship at Yidu Technology.}

\begin{abstract}
As large language models (LLMs) demonstrate increasing capability in generating research ideas, automating novelty evaluation has become a critical challenge for AI-driven scientific discovery. This paper presents Relative Neighbor Density (RND), a domain-agnostic algorithm for novelty assessment in research ideas that overcomes the limitations of existing approaches by comparing an idea's local density with its adjacent neighbors' densities. We first developed a scalable methodology to create test set without expert labeling, addressing a fundamental challenge in novelty assessment. Using these test sets, we demonstrate that our RND algorithm achieves state-of-the-art (SOTA) performance in computer science (AUROC=0.820) and biomedical research (AUROC=0.765) domains. Most significantly, while SOTA models like Sonnet-3.7 and existing metrics show domain-specific performance degradation, RND maintains consistent accuracies across domains by its domain-invariant property, outperforming all benchmarks by a substantial margin (0.795 v.s. 0.597) on cross-domain evaluation. These results validate RND as a generalizable solution for automated novelty assessment in scientific research.
\end{abstract}

% keywords can be removed
% \keywords{First keyword \and Second keyword \and More}

\section{Introduction} \label{sec: intro}
% Context 
As AI research advances toward AGI, automating scientific discovery processes becomes increasingly important. While large language models (LLMs) now demonstrate capability in generating research ideas, such as AI scientist (\cite{lu_ai_2024}), a critical challenge remains in how to pick pearls from those generated ideas, i.e., how to reliably evaluate these ideas for novelty. Traditionally, novelty in research ideas has been assessed through expert evaluation, where domain specialists judge originality based on their knowledge and experience. However, such assessments are inherently subjective, time-consuming, and inconsistent across evaluators. Automated methods for novelty assessment are therefore crucial for identifying truly innovative directions, particularly as AI systems begin to participate more actively in the research process.

% Existing approaches
Existing approaches primarily fall into two categories: (1) leveraging large language models (LLMs) as judges and (2) using absolute local density-based novelty metrics. LLM-based approaches include Swiss system tournaments (\cite{si_can_2024, hu_nova_2024}) and augmented judgment using NeurIPS review guidelines and Semantic Scholar API (\cite{lu_ai_2024, su_two_2024}). Alternatively, absolute local density metrics measure novelty through Euclidean distances between embeddings, as demonstrated by Su et al.'s Overall Novelty (ON) metric (\cite{su_two_2024}), which combines Historical Dissimilarity, Contemporary Dissimilarity, and citation-based Contemporary Impact, showing strong correlation with human-labeled novelty.

% Challenges
Despite these advances, significant challenges remain. First of all, in studies of any existing approach, the validation methodologies typically rely on small, domain-specific test sets with human-labeled data that quickly become outdated as research advances, limiting scalability and long-term accuracy. Additionally, the LLM-based judgment exhibits sensitivity to input perturbations (\cite{zhuo_robustness_2023, singh_robustness_2024}), potentially yielding inconsistent novelty ratings for semantically identical ideas. Finally, absolute local density metrics suffer from arbitrary parameter choices and lack generalizability across research domains with varying citation patterns and publication velocities. 

% How to address
To address these challenges, we establish comprehensive semantic embedding databases for novelty assessment, incorporating publications from two distinct domains: Pubmed, the leading biomedical literature search engine with nearly 36 million articles (\cite{jin2024pubmed}), and Arxiv, which contains more than 2.3 million scholarly articles across eight subject areas (\cite{arxiv2023report}). Based on these resources, we develop a methodology to create test sets with trustworthy novelty labels without requiring expert manual labeling. 

\begin{figure}
  \centering
  \includegraphics[width=\textwidth]{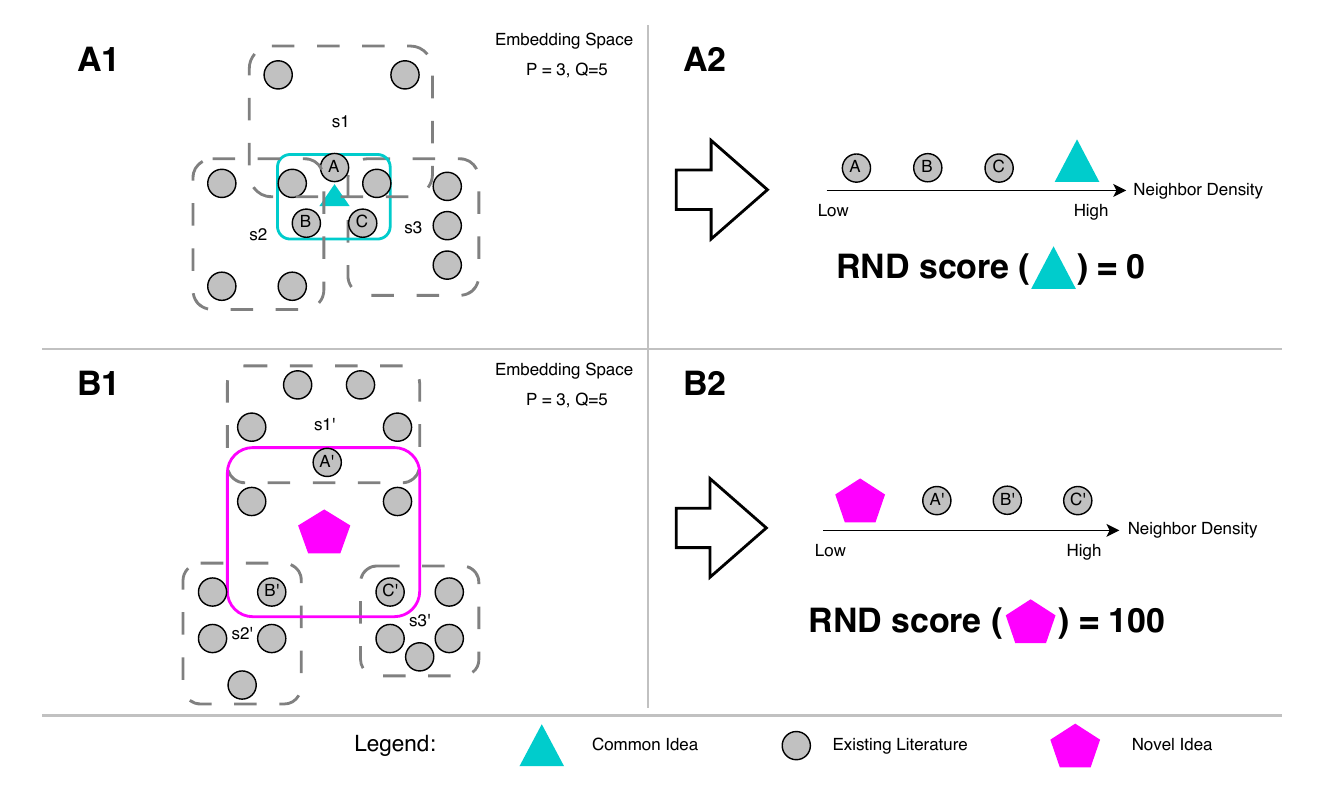}
  \caption{Illustration of the Relative Neighbor Density (RND) algorithm. 
  \textit{A1/B1}: In this step, both the given idea (triangle in A1 and pentagon in B1) and all existing literature in a given research domain are represented in a semantic embedding space. The $P$ nearest neighbors (A/B/C or A'/B'/C') of the given idea are identified. Then, for each of these neighbors, the neighbor density is computed by identifying $Q$ nearest surrounding neighbors (neighbor sets s1-s3 in A1 and s1'-s3' in B1). 
  \textit{A2/B2}: The neighbor densities of the $P$ closest pieces of literature and the given idea are sorted. The RND score of the given idea is determined based on its relative rank among these neighbor densities.}
  \label{fig: overall_algo}
\end{figure}

Furthermore, we propose the Relative Neighbor Density (RND) algorithm, which measures novelty by analyzing the distribution patterns of semantic neighbors rather than simple absolute local density (Figure \ref{fig: overall_algo}). This approach proves more reliable than LLM-based judgments and more generalizable than existing absolute local density-based metrics across different research domains. Our extensive evaluations using test sets from computer science, biomedical science, and cross-domain contexts demonstrate that our proposed algorithm maintains accuracy within specific domains while scaling effectively across diverse research areas.

Our main contributions are:
\begin{itemize}
\item A novel neighbor density-based Relative Neighbor Density (RND) algorithm for assessing research idea novelty that is robust across domains, which holds domain-invariant property
\item A scalable methodology for validating novelty metrics without expert labeling
\item Comprehensive evaluations comparing SOTA reasoning models, LLMs and algorithms for assessing novelty across multiple research domains
\end{itemize}

\section{Related Works} \label{sec: related}

\subsection{LLMs for Novelty Assessment}
Recent work has demonstrated promising results in using LLMs as autonomous judges for research novelty. Si et al. (\cite{si_can_2024}) evaluated this approach using ICLR submissions, converting them into standardized project proposals and conducting pairwise comparisons between accepted and rejected papers(Table \ref{tab:Standardized Project Proposals prompt}). Their Swiss tournament system iteratively paired proposals based on accumulated scores, with Claude-3.5-Sonnet achieving 71.4\% accuracy in predicting paper acceptance. As a control measure, they included human expert reranking, which revealed notable discrepancies between automated and human judgments.

Lu et al. (\cite{lu_ai_2024}) expanded this concept with their AI Scientist framework, integrating idea generation, evaluation, and refinement. Their system employs chain-of-thought prompting and external knowledge retrieval via Semantic Scholar API to enhance assessment quality. While showing promise in matching human-level performance, these LLM-based approaches face fundamental challenges in reliability and consistency, as highlighted by studies showing their sensitivity to input variations (\cite{zhuo_robustness_2023, singh_robustness_2024}).

\subsection{Absolute Local Density-based Metrics} \label{related: ON_HD_CI_CD}
An alternative approach focuses on semantic local density to evaluate novelty. Su et al. (\cite{su_two_2024}) used the Historical Dissimilarity (HD), which is the average Euclidean distance between the generated abstract embedding and embeddings of the 5 most similar abstracts in the historical literature base. We denote it as "Absolute Local Density" because the average distance is a metric of local density, and they use the value of density directly. 

In addition to HD, they also developed Overall Novelty (ON) as below, where CI is Contemporary Impact, the average citation count of the top 5 most similar abstracts in contemporary literature base; CD is Contemporary Dissimilarity, calculated by same algorithm in contemporary literature base.
\begin{equation}\label{eq:ON}
    ON = \frac{HD \times CI}{CD}
\end{equation}

The main challenge of local density-based metric is that the density values vary across different domains. In case of evaluating ideas from mixed research domains, the variance would cause a severe degrade in the accuracy.

\subsection{Validating Novelty Metrics}
A critical limitation in existing research is the lack of scalable validation methodologies. Current approaches (\cite{lu_ai_2024, su_two_2024, hu_nova_2024, si_can_2024}) typically rely on small-size literature database and manually labeled test set created by domain experts. For instance, Su et al. (\cite{su_two_2024}) constructed an "ecosystem" for computer science (CS) using information extracted from 85,217 papers—a dataset that represents only a small fraction of the CS literature available on platforms like arXiv. While their analysis demonstrated promising correlations between novelty scores and human labels across 100 manually evaluated abstracts, the methodology's reliance on domain-specific expertise significantly constrains its generalizability, which is echoed by reviewer's comments that only validating CS is "relatively simple" (\cite{two_head_validation_weakness}).

\section{Method} \label{sec: method}

\subsection{Problem Description}
Given a set of ideas $I$, \begin{equation}
    I = \{{idea}_i\}, i \in [1, N]
\end{equation}
Where ${idea}_i$ is a sequence of words or characters in nature language. $N \geq 1$ represents the number of ideas whose novelty needs to be assessed.

The objective is to design a mapping $\mathscr{F}$ from idea space to a score in real value space
\begin{equation}
    F({idea}_i) = {score}_i, \quad \text{where} \quad {idea}_i \in I, \, {score}_i \in \mathbb{R}
\end{equation}

The novelty score \( \text{score} \) should be \textbf{monotonic}, meaning that for any two ideas \( \text{idea}_i \) and \( \text{idea}_j \), if \( \text{idea}_i \) is more novel than \( \text{idea}_j \), then their corresponding scores must satisfy:  
\begin{equation}
    \forall \, \text{idea}_i, \text{idea}_j \in I, \quad \text{idea}_i \succ \text{idea}_j \Rightarrow F(\text{idea}_i) > F(\text{idea}_j)
\end{equation}
where \( \text{idea}_i \succ \text{idea}_j \) denotes that \( \text{idea}_i \) is considered more novel than \( \text{idea}_j \) based on a given novelty criterion.

\subsection{Semantic Embedding \& Literature Database}
Each published literature's abstract, which is also a sequence of words or characters in natural language, is denoted as \( a_j \).

The semantic embedding model is a mapping function \( \mathscr{G} \), which maps ideas and abstracts into embedding vectors:
\begin{align}\label{eq: convert_embedding}
    \mathscr{G}(\text{idea}_i) = \mathbf{v}_i, \quad \text{where} \quad \mathbf{v}_i \in \mathbb{R}^{\text{dims}}, \\
    \mathscr{G}(a_j) = \mathbf{v}_j, \quad \text{where} \quad \mathbf{v}_j \in \mathbb{R}^{\text{dims}}
\end{align}

Thus, the preprocessed literature semantic database is represented as a set \( A \):
\begin{equation}
    A = \{(a_j, \mathbf{v}_j) \mid j \in [1, M] \}
\end{equation}

We collected 36 million academic articles from the PubMed Download API (\cite{pubmed_download}) and 2.6 million papers from the ArXiv dataset (\cite{cornell_arxiv_dataset}). Among all fetched documents, only those with both a non-empty title and abstract were considered valid for the experiment, resulting in 25,360,114 papers from PubMed and 2,643,057 papers from ArXiv.

For each paper, two semantic embedding vectors were generated—one from its title and another from its abstract—using the M3-Embedding model~\cite{chen2024bgem3embeddingmultilingualmultifunctionality}. The embedding vector dimension, denoted as \( \text{$dims$} \), is 1024. All texts and embedding vectors were stored in Elasticsearch Version 8 for efficient retrieval.

\subsection{Algorithm}
For each idea \( \text{idea}_i \) and its embedding \( \mathbf{v}_i \), we first find its \( P \) nearest neighbors using \( k \)-Nearest Neighbors (KNN) search:
\begin{equation}\label{eq: 1st_level_neighbor}
    \{\mathbf{v}_{1}, \mathbf{v}_{2}, \dots, \mathbf{v}_{P}\} = KNN(\mathbf{v}_i, P, A)
\end{equation}
where \( \mathbf{v}_{j} \) is the \( j \)-th nearest neighbor of \( \mathbf{v}_i \), \( j \in [1, P] \).

For the idea itself \( \mathbf{v}_i \) and each neighbor \( \mathbf{v}_{j} \), the neighbor density (ND) is defined as below
\begin{equation}\label{eq: neighbor_density}
    ND = \frac{1}{Q}\sum^{Q}_{k=1}d(\mathbf{v}, \mathbf{v}_k)
\end{equation}
where $\mathbf{v}_k$ denotes the  \( k \)-th nearest article's embedding vector in the literature corpus \( A \) and \( d(\cdot , \cdot) \) is the cosine distance between two vectors.

We define the set \( S_i \) that contains the neighbor density values of \( \text{idea}_i \)'s neighbors:
\begin{equation}\label{eq: density_list}
    S_i = \{ND_{j} \mid j \in [1, P]\}
\end{equation}

Finally, we compute the novelty score \( \text{score}_i \) for \( \text{idea}_i \) as:
\begin{equation}\label{eq: score}
    \text{score}_i = \frac{\left| \{ ND \in S_i \mid ND \leq ND_i \} \right|}{|S_i|} \times 100
\end{equation}

In Appendix \ref{appendix: alg}, the pseudocode for ND calculation is presented in Algorithm~\ref{alg:get_neighbors_and_density}; and the pseudocode of complete algorithm is provided in Algorithm~\ref{alg: check if an idea is novel}.

In this algorithm, the selection of values of \( P \) and \( Q \) is a trade-off between reliability of estimation and other cost.  Refer to Appendix \ref{appendix: alg} for mathematical analysis and Section \ref{sec: sen_hyper_param} for empirical experimental results. Based on analysis and empirical experiments, we set \( P = 100 \) and \( Q = 50 \). 

\subsection{Validation without Human Labeling} \label{sec: method_val}
As a novelty evaluation algorithm, the most challenging point in past research is to find a reliable labeled test set to evaluate the algorithm. Therefore, we propose a new method to construct a convincing test set instead of relying on human experts to annotate it.

For the positive samples (a.k.a novel ideas) in the test set, we select recent articles from top journals or conferences. For the negative samples (a.k.a. non-novel ideas), highly cited articles published before the last few years were selected, also from the research domain's top journals or conferences. The fundamental principles behind such methodology were: high-quality novel ideas are more likely to be published in recent issues and top journals or conferences; while after time passes, the at-the-time novel ideas were more likely to attract attention and related works, thus become non-innovative at present. 

\section{Result} \label{sec: result}

\subsection{Experiment Setup} \label{sec: result_exp_setup}
\subsubsection{Test Set}
We have two test sets: NeurIPS, which represents the most advanced research results in the field of computer science, and Nature Medicine, which represents the most cutting-edge papers in the medical field. The sample year distribution of the test sets can be found in Table \ref{tab:Count of Data in Different Time Ranges}

\textbf{NeurIPS test set}: The initial corpus consists of papers that are Accept (oral) or Accept (spotlight) by Program Chairs at the 2024 NeurIPS conference, which represents the latest research results in computer science. Furthermore, we select articles from the initial corpus that explicitly mention that the papers have obvious novelty in the comments of Program Chairs to form the positive samples of the NeurIPS test set. The comments and decision information of Program Chairs can be obtained on the OpenReview.net website. At the same time, we use the Semantic Scholar API to obtain the 99 most cited papers published in the NeurIPS conference from 2015 to 2020 to form the negative samples of the test set. The titles of all samples are presented in Table \ref{tab:NIPS_Dataset}

\textbf{Nature Medicine test set}: The positive samples of the Nature Medicine test set consist of articles classified as "Article" type, published in Nature Medicine from August 2024 to February 2025, according to the classification on the nature.com website. Articles related to phase 2 or phase 3 trials were excluded. And we used the same method as the negative samples of the NeurIPS test set to obtain 99 articles of Nature Medicine with the highest citation count in the past 15 years as negative samples of the test set. The titles of all samples are presented in Table \ref{tab:Nature_Medicine_Dataset}

\textbf{Mixed Test set}: The mixed test set is the combination of the negative samples from NeurIPS and positive samples from Nature Medicine.

\begin{table}
\centering
\caption{Count of Data in Different Time Ranges for NeurIPS and Nature Medicine Test Sets. \textit{Positive}: novel samples, \textit{Negative}: non-novel samples.}
\begin{tabular}{@{}ccccccc@{}}  % 使用 @{} 来消除表格两边的额外空白
\toprule  % 表头上方的横线
\multirow{2}{*}{\textbf{Test set}} & \multirow{2}{*}{\textbf{Label}} & \multicolumn{5}{c}{\textbf{Count}} \\  % 第一行
\cmidrule(lr){3-7}  % 表头下面的横线，划分年份列
                                  & & \textbf{total} & \textbf{2024-2025} & \textbf{2019-2023} & \textbf{2014-2018} & \textbf{-2014} \\
\midrule  % 表头下方的横线
\multirow{2}{*}{NeurIPS}          & Positive       & 80 & 80 & 0  & 0  & 0  \\
                                  & Negative       & 99 & 0  & 31 & 68 & 0  \\
\midrule
\multirow{2}{*}{Nature Medicine}  & Positive       & 66 & 66 & 0  & 0  & 0  \\
                                  & Negative       & 99 & 0  & 29 & 32 & 38 \\
\bottomrule  % 表格底部的横线
\end{tabular}
\label{tab:Count of Data in Different Time Ranges}
\end{table}

\subsubsection{Baseline}
To evaluate our algorithm, we selected all existing novelty assessment algorithms as baselines, categorized into two groups: LLM-based and non-LLM-based. Non-LLM-based algorithms, including Relative Neighbor Density(Ours), Historical Dissimilarity(HD), and Overall Novelty(ON), rely solely on literature search and mathematical calculations. Since the output of the literature search for the same query remains consistent, we conducted a single test to assess the algorithm's performance. In contrast, for LLM-based algorithms, due to the inherent variability of LLM outputs, we ran three tests for each algorithm, calculated the average result, and included the standard deviation in the results.

For all algorithms tested, we use the abstracts of the papers as "ideas" for testing.

\textbf{Historical Dissimilarity (HD)} \& \textbf{Overall Novelty (ON)}: Both HD and ON are described in Section \ref{related: ON_HD_CI_CD}. During calculation, the historical database contains papers from 2011 to 2021, and the contemporary database contains papers from 2021 to 2025. 

\textbf{LLM + literature search}: Provide LLM with the titles and abstracts of the 10 most relevant papers to the given idea. The model then assesses whether the core concepts of these papers significantly overlap with the idea(Table \ref{tab:LLM_with_paper_prompt}). Sonnet-3.7 (\cite{anthropic_2025_sonnect_37}), Deepseek-r1 (\cite{guo2025deepseek_r1}) and GPT-4o (\cite{hurst2024gpt4o}) were selected as the LLMs to be tested.

\textbf{LLM with guideline}: Utilize the NeurIPS 2024 review guidelines to assist LLM in evaluating the novelty of ideas(Table \ref{tab:LLM_with_guideline_prompt}). Sonnet-3.7 is selected as the LLM to be tested.

\textbf{LLM with tournament}: First, the idea is transformed into the Standardized Project Proposal format(Table \ref{tab:Standardized Project Proposals prompt}). Next, the novelty of all standardized ideas is assessed using the Swiss tournament method, where ideas are iteratively compared in a structured competition. Finally, each idea is assigned a score based on the number of wins it accumulates throughout the tournament. Sonnet-3.7 is selected as the LLM to be tested.

\subsection{Accuracy Evaluation}

\begin{table}
\centering
\caption{Validation of Different methods, measured by AUROC. 
\textit{HD}: Historical Dissimilarity (section \ref{related: ON_HD_CI_CD}). 
\textit{ON}: Overall Novelty (section \ref{related: ON_HD_CI_CD}). 
\textit{LLM + literature search}: supplementing LLM with 10 relevant papers, which were searched by idea's embedding from our literature database using semantic embedding. 
\textit{LLM with guideline}: using NeurIPS 2024 review guideline to help LLM judge the novelty of ideas, which is not applicable to Nature Medicine. Therefore, the results of Nature Medicine and Mixed are marked as not applicable.
\textit{LLM with tournament}: a Swiss system tournament design to evaluate ideas by using LLM as judge. 
}
\begin{tabular}{@{}c|p{2cm}ccc@{}}  % 使用 @{} 来消除表格两边的额外空白
\toprule  % 表头上方的横线
\multicolumn{2}{c}{\textbf{Model}}& \textbf{NeurIPS} & \textbf{Nature Medicine} & \textbf{Mixed} \\

\midrule  % 表头下方的横线
\multicolumn{2}{c}{\textbf{\textit{Relative Neighbor Density}(Ours)}} &  0.820 & \textbf{0.765} & \textbf{0.795} \\
\cmidrule(lr){1-2} % 仅在前两列添加分割横线
\multirow{2}{*}{\textbf{Absolute Local Density} }& \textbf{HD} & \textbf{0.856} & 0.699 & 0.362 \\
& \textbf{ON} & 0.584 & 0.544 & 0.456 \\
\cmidrule(lr){1-2} % 仅在前两列添加分割横线
\multirow{3}{*}{\textbf{LLM + literature search}}& \textbf{Sonnet-3.7} & $0.813\pm0.01$ & $0.616\pm0.006$ & $0.597\pm0.004$ \\
& \textbf{Deepseek-r1} & $0.710\pm0.027$ & $0.673\pm0.025$ & $0.596\pm0.049$ \\
& \textbf{GPT-4o} & $0.567\pm0.008$ & $0.545\pm0.02$ & $0.522\pm0.022$ \\
\cmidrule(lr){1-2} % 仅在前两列添加分割横线
\multicolumn{2}{c}{\textbf{Sonnet-3.7 with guideline}} & $0.546\pm0.035$ & NaN & NaN \\
\multicolumn{2}{c}{\textbf{Sonnet-3.7 with tournament}} & $0.496\pm0.001$ & $0.503\pm0.005$ & $0.501\pm0.004$ \\
\bottomrule  % 表格底部的横线
\end{tabular}
\label{tab: AUROC Comparison of Different Models}
\end{table}

As shown in Table \ref{tab: AUROC Comparison of Different Models}, our enhanced neighbor density-based novelty measurement algorithm outperforms all baseline models on the Nature Medicine and Mixed test sets, and performs strongly on the NeurIPS test set.

By comparing the results of various LLM-related algorithms, we find that both \textbf{Sonnet-3.7 with guideline} and \textbf{Sonnet-3.7 with tournament} show limited external knowledge input, leading to inaccurate novelty judgments. In contrast, the \textbf{LLM + literature search} method, which incorporates the 10 most relevant papers based on semantic embedding, significantly improves accuracy. This method shows a higher AUROC in computer science than in biomedicine, highlighting the impact of the model’s internal knowledge on judgment outcomes.

Additionally, Sonnet-3.7 and Deepseek-r1 outperform GPT-4o, suggesting that external knowledge enhances the performance of the inference model compared to autoregressive models.

We also observed that the Historical Dissimilarity (HD) metric performed similarly to our method on the Nature Medicine and NeurIPS test sets. However, on the Mixed test set, there was a significant disparity, with our method achieving an AUROC of 0.795, while HD reached only 0.362. This difference reflects the limited generalization ability of HD across domains. In contrast, the distributions of our method’s scores are consistent across all test sets, indicating its robust cross-domain evaluation capability, making it applicable to various research fields.

\begin{figure}
  \centering
  \includegraphics[width=\textwidth]{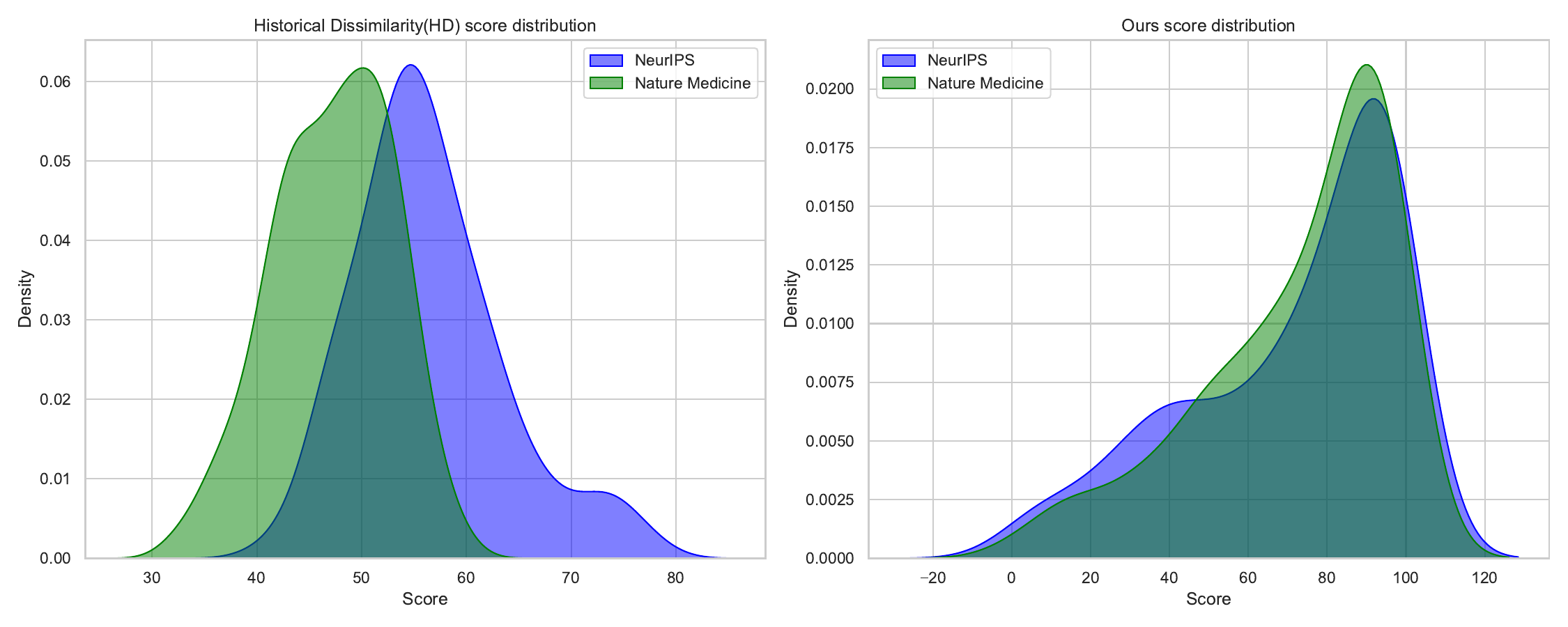}
  \caption{Comparison of HD \& Our score distributions in different domains. \textit{1}: In the right panel, the upper and lower bounds of the score exceeded the actual score range ($[0, 100]$) because of linear interpolation. \textit{2}: to make the horizontal axis comparable, we scaled the Historical Dissimilarity scores by $\times 100$.}
  \label{fig: Comparison of score distributions}
\end{figure}

\subsection{Sensitivity Study} \label{sec: sen_study}
\subsubsection{Sensitivity of Hyper-Parameter} \label{sec: sen_hyper_param}

\begin{figure}
  \centering
  \includegraphics[width=\textwidth]{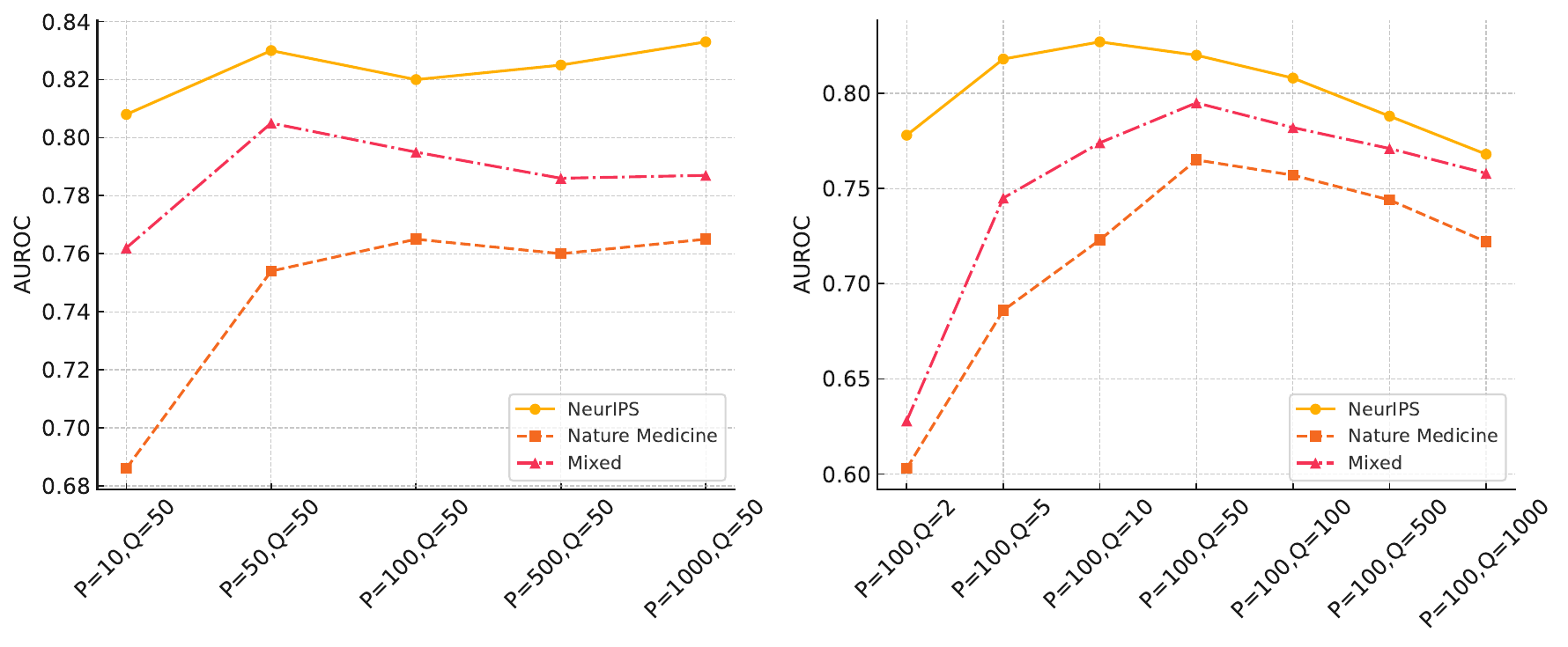}
  \caption{\textbf{Comparison of AUROC of RND algorithm with different parameters}. \textit{left}: AUROC with different P value when Q=50. \textit{right}: AUROC with different Q value when P=100}
  \label{fig: AUROC_Comparison_P_Q}
\end{figure}

As illustrated in the left panel of Figure~\ref{fig: AUROC_Comparison_P_Q}, the AUROC on each test set increases as $P$ grows. However, when $P > 50$, the improvement in AUROC becomes marginal compared to the significant gain observed when increasing $P$ from 10 to 50. This suggests that the marginal benefit of further increasing $P$ diminishes while simultaneously incurring substantial computational costs, given that the algorithm's time complexity is $O(P \cdot Q)$. Additionally, the poor performance observed when $P = 10$ can be attributed to the biased estimation of novelty scores when $P$ is too small, a phenomenon influenced by multiple factors. For a more detailed explanation, please refer to Appendix \ref{appendix :Effects of P}.

The right panel of Figure~\ref{fig: AUROC_Comparison_P_Q} demonstrates that when $P$ remains constant, both excessively small and large values of $Q$ negatively impact the algorithm's performance. This is due to the inaccuracy in local density estimation when $Q$ is too small and the significant reduction in algorithm sensitivity when $Q$ is too large. For further details, please refer to Appendix~\ref{appendix: Effects of Q}.

\subsubsection{Sensitivity of Design}

In our Relative Neighbor Density algorithm, the notion "relative", i.e. comparing idea's local density with its neighbor's local densities, plays an important role. Moreover, other distance metric, such as Euclidean distance, could also be used in our algorithm. To understand the sensitivity of the current design, we conducted experiments by changing the design of the "relative" notion, and distance metric. The result presented in Table \ref{tab: AUROC Comparison between Ours and Different Ablations} validate our statement.

\begin{table}
\centering
\caption{AUROC Comparison for Different Design . \textit{Absolute Local Density}: Use the idea's local density as novelty score. (density calculated by mean distances between idea and idea's \( P \) first level neighbors). \textit{Euclidean distance}: replace the cosine distance with Euclidean in RND}
\begin{tabular}{@{}cccc@{}}  % 使用 @{} 来消除表格两边的额外空白
\toprule  % 表头上方的横线
\multirow{2}{*}{\textbf{Test set}} & \multicolumn{3}{c}{\textbf{AUROC}} \\  % 第一行
\cmidrule(lr){2-4}  % 表头下面的横线，划分年份列
                                  & \textbf{Ours}& \textbf{Absolute Local Density} & \textbf{Euclidean distance}\\
\midrule  % 表头下方的横线
NeurIPS         &  0.820          & \textbf{0.851 }      &  0.815\\
Nature Medicine & \textbf{0.765}            & 0.757       & 0.753\\
Mixed  & \textbf{0.795}         & 0.395       & 0.78\\
\bottomrule  % 表格底部的横线
\end{tabular}
\label{tab: AUROC Comparison between Ours and Different Ablations}
\end{table}

\begin{figure}[h]
  \centering
  \includegraphics[width=\textwidth]{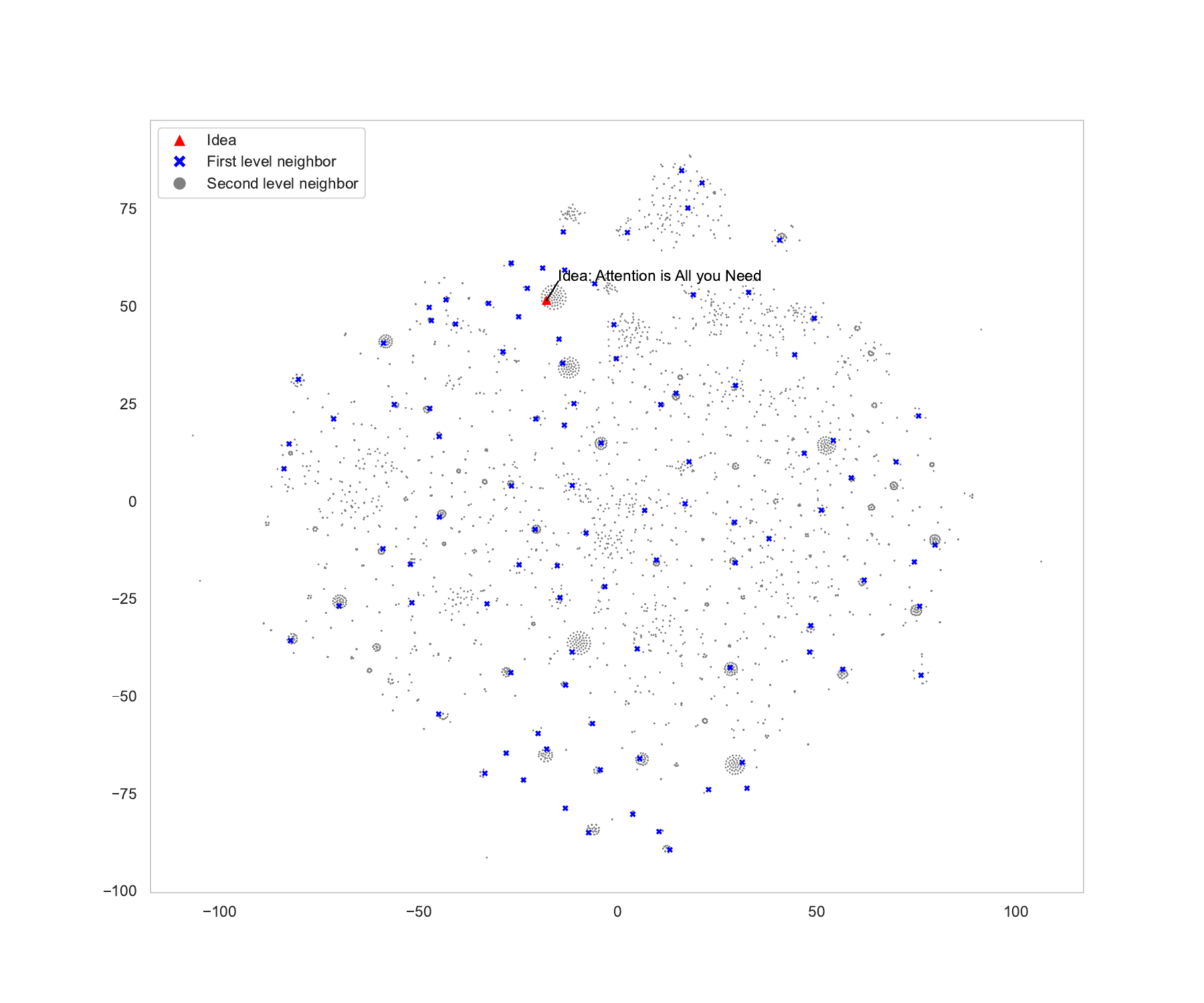}
  \caption{\textbf{Neighbor Distribution of a Non-novel Idea in Embedding Space (t-SNE processed)}.}
  \label{fig: Neighbor Distribution of a Non-novel Idea}
\end{figure}

\begin{figure}[H]
  \centering
  \includegraphics[width=\textwidth]{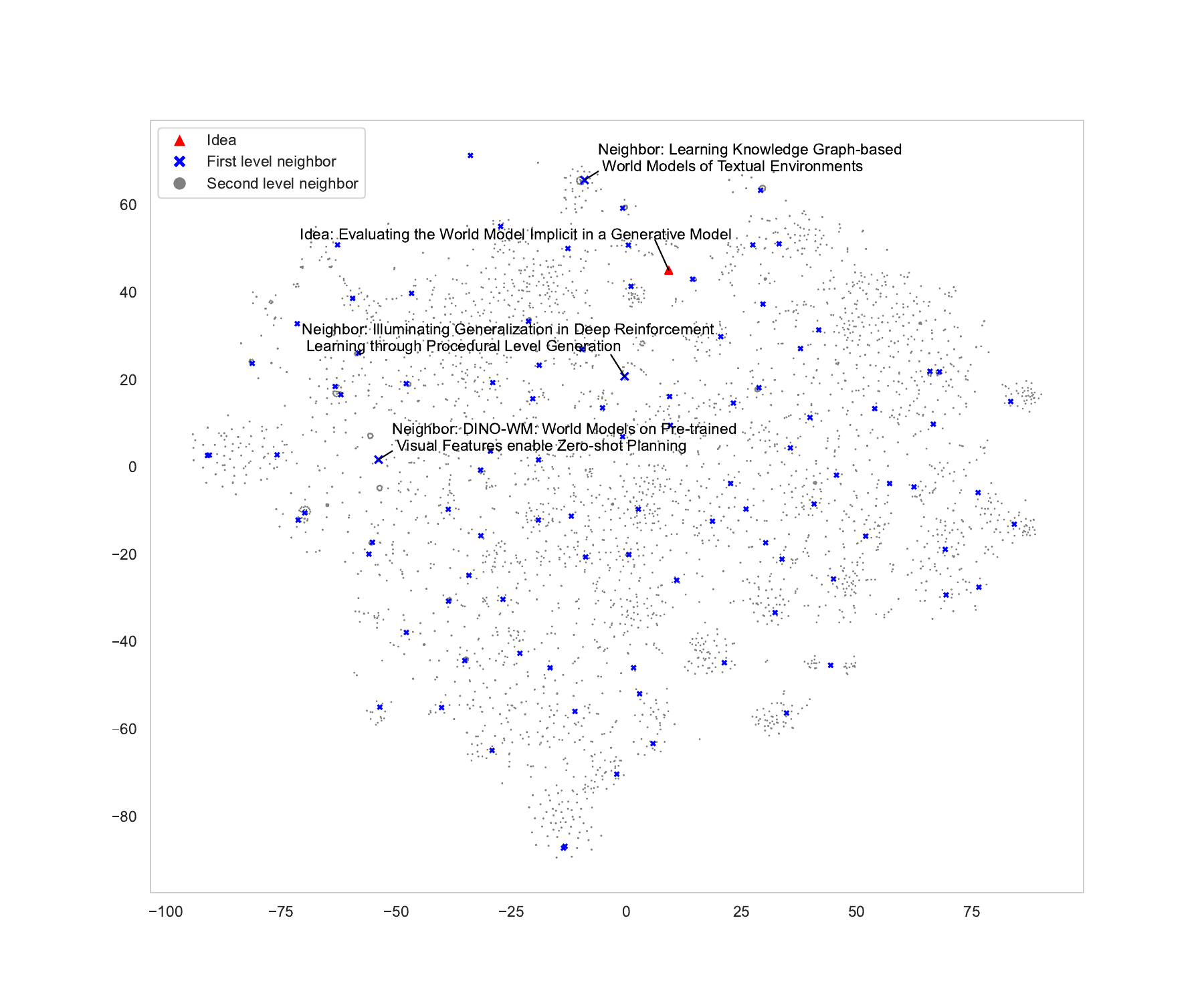}
  \caption{\textbf{Neighbor Distribution of a Novel Idea in Embedding Space (t-SNE processed)}.}
  \label{fig: Neighbor Distribution of a Novel Idea}
\end{figure}

\subsection{Case Study}
To visualize the geometric structure of ideas and their neighboring points in the embedding space, we selected two ideas and plotted their surrounding neighbors on a 2-D plane using t-SNE (\cite{van2008visualizing}).

We first examine \textit{Attention is All You Need} (\cite{vaswani2017attention}), a highly cited article, as an example of a non-novel idea from the current perspective. As shown in Figure \ref{fig: Neighbor Distribution of a Non-novel Idea}, there is a dense cluster of neighbors surrounding the idea, while the neighbors around its \( P \) nearest neighbors are relatively fewer and more sparsely distributed. 

Next, we analyze \textit{Evaluating the World Model Implicit in a Generative Model} (\cite{vafa2025evaluating}), an article recognized as highly novel by the NeurIPS 2024 Program Chairs based on their comments (\cite{openreview_2025_Evaluating-the-World-Model}). In Figure \ref{fig: Neighbor Distribution of a Novel Idea}, the idea’s local neighborhood appears considerably sparser than that of its \( P \) nearest neighbors.

Moreover, the differences between the two figures extend beyond the local density around the central idea; the local densities of neighboring ideas also exhibit distinct patterns. In Figure \ref{fig: Neighbor Distribution of a Non-novel Idea}, multiple neighboring clusters centered around the idea’s \( P \) nearest neighbors are visible, whereas such clustering patterns are absent in the novel idea’s neighborhood. These observations suggest that an idea’s novelty is associated not only with its own local density but also with the density distribution of its neighboring ideas.

\section{Discussion} \label{sec: discussion}
% 0. Brief Summary
In this work, we proposed a novel neighbor density-based metric for assessing research idea novelty, addressing the limitations of LLM judgment and absolute local density-based metrics. By leveraging large-scale literature embeddings from both biomedical sciences and computer science, our approach ensures robust reliability and cross-domain generalizability. Additionally, we introduced a scalable validation framework that eliminates reliance on expert labeling, enabling objective and reproducible novelty assessment.

\subsection{Why a Non-LLM Novelty Assessment Algorithm is Necessary?}

While LLMs have the potential to assess novelty, the reliability of their judgment is limited, as outlined in the Introduction section (\ref{sec: intro}). Our experiments (see Table \ref{tab: AUROC Comparison of Different Models}) echoed such concern: without an integrated search tool, even the most advanced reasoning models' performance was comparable to random guessing (AUROC\~=0.5). When a search tool was introduced, Sonnet-3.7 achieved similar accuracy on the NeurIPS test set (AUROC\~=0.8) but experienced significant degradation (AUROC\~=0.6) on both the Nature Medicine and cross-domain test sets.

In contrast, our proposed RND algorithm can produce more reliable and consistent results, as seen in Table \ref{tab: AUROC Comparison of Different Models}. Additionally, our algorithm is better at distinguishing genuinely novel ideas from the large pool of candidates from mixing research domains (AUROC\~=0.78 v.s Other's AUROC<=0.6). Such cross-domain novelty assessment capability is crucial to AI scientist, as more and more innovation happens in the intersection of research domains.

\subsection{Domain-invariant Accuracy}

The most significant advantage of RND algorithm is in being able to accurately compare novelty across different domains.

As demonstrated in the Table \ref{tab: AUROC Comparison between Ours and Different Ablations}, though local density-based HD algorithm and LLM with literature search tools achieved 0.8+ AUROC in NeruIPS test set, their performance degraded in Nature Medicine test set (0.699 for HD, 0.673 for Deepseek-r1). Moreover, when tested in the cross-domain test set (the Mixed test set), the AUROC of HD and LLM significantly degraded below 0.6. In contrast, the AUROC of our proposed RND algorithm remained robust (0.820, 0.765, 0.795) across each single domain and mixed domain test sets. 

We argue that such benefits come from RND algorithm's \textbf{domain-invariant property}, i.e. the distribution of novelty scores produced by RND is identical regardless of the tested domain, which explained why our relative density-based approach succeeds in cross-domain scenarios. According to the mathematical reasoning in Appendix \ref{appdix: invariant}, we concluded the distribution of novelty score \( S \) is only subject to \( P \) (the number of neighbors considered); thus it is invariant to the validation domain. Furthermore, in Figure \ref{fig: Comparison of score distributions} (right panel) and \ref{fig: Score distribution with dif P}, the actual distribution of scores echoed the theoretical analysis. Such domain-invariant property is crucial for conducting multi-disciplinary scientific research, where ideas from diverse fields must be compared and evaluated effectively.

\subsection{Limitations \& Future Work}

First, the RND algorithm's effectiveness is inherently linked to the breadth and quality of the underlying literature database and semantic embedding model. Incomplete literature coverage or suboptimal embedding quality can impact novelty assessment accuracy. While implementing RND requires more infrastructure than LLM-as-judge approaches, this investment enables consistent, deterministic assessments that remain stable across model versions and prompting strategies.

Second, in the test set created by our validation methodology, while the need for expert labeling was avoided, the non-novel samples may be too easily distinguishable from novel ones. By using historical highly-cited papers as non-novel examples, rather than borderline cases such as recently rejected papers or incremental work from current journals, we created a simpler assessment scenario than the actual scenario in real research settings. However, the fact that none of the tested algorithms achieved saturated AUROCs even in this relatively straightforward scenario demonstrates the fundamental challenge of novelty assessment.

While acknowledging these limitations, we believe our work establishes a solid foundation for several promising research directions that could significantly advance automated scientific discovery.

1. Integration with AI Research Workflows: Incorporating our novelty evaluation algorithm into end-to-end AI scientist workflows would enable autonomous research ideation and evaluation. This integration would allow AI systems to independently generate research hypotheses, assess their novelty using our domain-invariant RND algorithm, and prioritize the most promising directions for further investigation. Such integration could accelerate scientific discovery by efficiently navigating complex multi-disciplinary research landscapes where human intuition about novelty is often limited. 

2. Enhancing Reasoning Capability for LLM: We propose utilizing our RND algorithm as a sophisticated reward mechanism within reinforcement learning frameworks for training reasoning models for AI research. By providing domain-invariant novelty signals during training, it could potentially guide models to generate more innovative scientific ideas while maintaining scientific validity.

\newpage
\bibliographystyle{plainnat}\bibliography{ref}

\newpage
\section*{\Large Appendix}
\appendix
\section{Algorithms} \label{appendix: alg}

\subsection{Pseudocode}
\begin{algorithm}
\caption{Find Neighbors and Calculate Neighbor Density}
\label{alg:get_neighbors_and_density}
\begin{algorithmic}[1]
\Function{Neighbor}{Input, P, Q}
    \State $\mathbf{v}_{Input} \gets$ \Call{Get\_Embedding}{Input} \Comment{Using M3-Embedding model}
    \State \textit{C} $\gets$ [] 
    \State \textit{neighbors} $\gets$ \Call{Get\_Neighbors}{$\mathbf{v}_{Input}$, $\max(P,Q)$} \Comment{Find $\max(P,Q)$ nearest neighbors}
    \State $neighbors\_for\_count \gets neighbors[:Q]$ \Comment{Only use the Q nearest neighbors to calculate density}
    \State $neighbors\_for\_distribution \gets neighbors[:P]$ \Comment{The P nearest neighbors are used to calculate distribution}
    \For{each paper in $neighbors\_for\_count$}
        \State $\mathbf{v}_{paper} \gets$ \Call{Get\_Embedding}{paper}
        \State \textit{distance} $\gets$ \Call{1 - cosine\_similarity}{$\mathbf{v}_{Input}, \mathbf{v}_{paper}$}
        \State \textit{C}.Append(\textit{distance})
    \EndFor
    \State $ND_{Input} \gets$ \Call{mean}{\textit{C}}
    \State \textbf{return} $ND_{Input}$, $neighbors\_for\_distribution$
\EndFunction
\end{algorithmic}
\end{algorithm}

\begin{algorithm}
\caption{Calculate Novelty Score of Given Idea}
\label{alg: check if an idea is novel}
\begin{algorithmic}[1]
\State \textbf{Input:} Idea
\State \textbf{Output:} A score in the range of 0 to 100
\State \textit{D} $\gets$ [] 
\State $ND_{Idea}$, \textit{neighbors} $\gets$ \Call{Neighbor}{Idea, P, Q}
\For {paper in \textit{neighbors}}
    \State $ND_{paper}$, \_ $\gets$ \Call{Neighbor}{paper, P, Q}
    \State \textit{D}.Append($ND_{paper}$)
\EndFor
\State \textit{score} $\gets \frac{| \{ ND \in D \mid ND \leq ND_{Idea} \} |}{|D|} \times 100$
\State \textbf{Return} \textit{score}
\end{algorithmic}
\end{algorithm}

\subsection{Effects of Parameter P}\label{appendix :Effects of P}

The novelty score is computed as
\begin{equation}\label{eq:score}
    \text{score}_i = \frac{\left| \{ ND \in S_i \mid ND \leq ND_i \} \right|}{P} \times 100,
\end{equation}
where \( S_i = \{ND_j \mid j = 1, 2, \dots, P\} \) is the set of neighbor densities for the \( P \) nearest neighbors of the idea \( i \), and \( ND_i \) is the neighbor density for idea \( i \) itself.

\subsubsection*{1. Empirical Cumulative Distribution Function (ECDF) Interpretation}
Define the empirical cumulative distribution function (ECDF) for the set \( S_i \) as
\[
F_P(x) = \frac{1}{P} \sum_{j=1}^{P} \mathbf{1}_{\{ ND_j \leq x \}},
\]
where \(\mathbf{1}_{\{ ND_j \leq x \}}\) is the indicator function that is 1 if \( ND_j \leq x \) and 0 otherwise. Then, by definition, the novelty score can be written as
\begin{equation}\label{eq:score_ecdf}
    \text{score}_i = 100 \cdot F_P(ND_i).
\end{equation}

\subsubsection*{2. Consistency of the ECDF}
Let \( F(x) \) be the true cumulative distribution function of the neighbor densities (assumed to be i.i.d. samples from a distribution \( F \)). By the Glivenko-Cantelli theorem, the ECDF \( F_P(x) \) converges uniformly to \( F(x) \) as \( P \to \infty \):
\[
\sup_{x} \left| F_P(x) - F(x) \right| \xrightarrow{P \to \infty} 0.
\]
Thus, for a sufficiently large \( P \), we have
\begin{equation}\label{eq:consistency}
    F_P(ND_i) \approx F(ND_i).
\end{equation}
This shows that the score, being proportional to \( F_P(ND_i) \), converges to \( 100 \cdot F(ND_i) \), which is the true quantile of \( ND_i \) in the distribution of neighbor densities.

\subsubsection*{3. Variance and Sensitivity with Finite \( P \)}
For a finite sample size \( P \), \( F_P(ND_i) \) is a random variable whose variance depends on \( P \). Under the assumption of i.i.d. sampling,
\[
\operatorname{Var}(F_P(ND_i)) = \frac{F(ND_i)(1 - F(ND_i))}{P}.
\]
Thus, the standard deviation is proportional to \( \frac{1}{\sqrt{P}} \). This quantifies that:
\begin{itemize}
    \item \(\textbf{Smaller } P:\) The variance \( \operatorname{Var}(F_P(ND_i)) \) is larger, leading to a noisier (less reliable) estimation of the quantile, and hence of the novelty score.
    \item \(\textbf{Larger } P:\) The variance decreases, yielding a more accurate estimation of the true quantile \( F(ND_i) \).
\end{itemize}
The novelty score becomes less sensitive to random fluctuations when \( P \) is large, as the empirical quantile is a better estimator of the true quantile.

\subsubsection*{4. Discreteness of the Score for Small \( P \)}
When \( P \) is small, the possible values of \( F_P(ND_i) \) are discrete, specifically:
\[
F_P(ND_i) \in \left\{ 0, \frac{1}{P}, \frac{2}{P}, \dots, 1 \right\}.
\]
For instance, if \( P=1 \), then \( F_1(ND_i) \) can only be 0 or 1, corresponding to a score of either \( 0\% \) or \( 100\% \). This coarse granularity can result in a biased or uninformative measure of novelty. As \( P \) increases, the steps \( \frac{1}{P} \) become finer, allowing the score to capture more subtle differences in the density distribution.

\subsubsection*{Conclusion}
The parameter \( P \) affects the final novelty score in two major ways:
\begin{enumerate}
    \item \textbf{Accuracy:} As \( P \) increases, the empirical cumulative distribution \( F_P(x) \) better approximates the true cumulative distribution \( F(x) \), leading to a more accurate quantile estimate \( F(ND_i) \).
    \item \textbf{Variance:} The variance of the estimate \( F_P(ND_i) \) is proportional to \( \frac{1}{P} \). Thus, a larger \( P \) reduces the variability of the score, making it less sensitive to random noise.
\end{enumerate}
In summary, a higher \( P \) leads to a more robust and sensitive measure of novelty, while a smaller \( P \) results in a discrete and noisier estimate.

\subsection{Effects of Parameter Q}\label{appendix: Effects of Q}

The neighbor density (ND) is given by:

\begin{equation}
    ND = \frac{1}{Q} \sum_{k=1}^{Q} d_k,
\end{equation}

where \( d_k = d(\mathbf{v}, \mathbf{v}_k) \) represents the distance between the point \( \mathbf{v} \) and its \( k \)-th nearest neighbor.

Assuming that \( d_k \) are independent and identically distributed (i.i.d.) random variables with mean \( \mathbb{E}[d_k] = \mu_d \), we compute the expectation of ND:

\begin{equation}
    \mathbb{E}[ND] = \mathbb{E} \left[ \frac{1}{Q} \sum_{k=1}^{Q} d_k \right] = \frac{1}{Q} \sum_{k=1}^{Q} \mathbb{E}[d_k] = \frac{1}{Q} Q \mu_d = \mu_d.
\end{equation}

The variance of ND is given by:

\begin{equation}
    \text{Var}(ND) = \text{Var} \left( \frac{1}{Q} \sum_{k=1}^{Q} d_k \right).
\end{equation}

Using the property that the variance of the mean of \( Q \) i.i.d. random variables is:

\begin{equation}
    \text{Var} \left( \frac{1}{Q} \sum_{k=1}^{Q} d_k \right) = \frac{1}{Q^2} \sum_{k=1}^{Q} \text{Var}(d_k).
\end{equation}

Since each \( d_k \) has variance \( \sigma_d^2 \), we obtain:

\begin{equation}
    \text{Var}(ND) = \frac{Q \sigma_d^2}{Q^2} = \frac{\sigma_d^2}{Q}.
\end{equation}

\subsubsection{Interpretation of Variance Scaling}

The derived formula:

\begin{equation}
    \text{Var}(ND) = \frac{\sigma_d^2}{Q}
\end{equation}

shows that:

\begin{itemize}
    \item As \( Q \) increases, the variance of ND decreases.
    \item Specifically, variance scales inversely with \( Q \), meaning that larger \( Q \) results in a more stable estimate of ND.
    \item When \( Q \to \infty \), \( \text{Var}(ND) \to 0 \), indicating that ND converges to its expected value \( \mu_d \), which would cause lost of information on local density.
    \item For small \( Q \), ND exhibits higher variability, making it more sensitive to local fluctuations.
\end{itemize}

\subsection{Domain-Invariant} \label{appdix: invariant}

\subsubsection{Theoretical Analysis}
Consider a test set in which each idea is assigned a neighborhood density defined as
\begin{equation}
ND = \frac{1}{Q}\sum_{i=1}^{Q} d(\text{idea}, a_i),
\end{equation}
where \(a_i\) denotes the \(i\)th nearest article in the literature corpus and \(d(\cdot,\cdot)\) is the cosine distance.

Let \(F(x)\) be the cumulative distribution function (CDF) of the neighborhood densities in the literature corpus. The percentile score for an idea is then defined by
\begin{equation}
S = F(ND).
\end{equation}
By the probability integral transform, if \(ND\) is drawn from a distribution with CDF \(F(x)\), then
\begin{equation}
S \sim \mathcal{U}(0,1).
\end{equation}

In practice, \(F(x)\) is estimated empirically using \(P\) articles from the neighborhood. The empirical CDF is given by
\begin{equation}
\hat{F}_P(x) = \frac{1}{P} \sum_{j=1}^{P} \mathbf{1}\{ND_j \leq x\},
\end{equation}
where \(ND_j\) is the neighborhood density of the \(j\)th article, and \(\mathbf{1}\{\cdot\}\) is the indicator function. Since
\begin{equation}
\sum_{j=1}^{P} \mathbf{1}\{ND_j \leq x\} \sim \operatorname{Binomial}(P, F(x)),
\end{equation}

we have
\begin{equation}\label{eq: mean_var}
\mathbb{E}[\hat{F}_P(x)] = F(x) \quad \text{and} \quad \operatorname{Var}[\hat{F}_P(x)] = \frac{F(x)(1-F(x))}{P}.
\end{equation}

Now, consider two literature corpora: a medical corpus with density distribution \(F_M(x)\) and a computer science corpus with density distribution \(F_C(x)\). For an idea in the test set, define its scores as
\begin{equation}
\hat{S}_M = \hat{F}_M(ND_M) \quad \text{and} \quad \hat{S}_C = \hat{F}_C(ND_C),
\end{equation}
where \(ND_M\) and \(ND_C\) are the neighborhood densities computed using the respective corpora. According to equation \ref{eq: mean_var}, we have
\begin{equation}
\mathbb{E}[\hat{S}] = F(ND) \quad \text{and} \quad \operatorname{Var}[\hat{S}] = \frac{F(ND)(1-F(ND))}{P}.
\end{equation}
where $F(ND) \sim \mathcal{U}(0,1)$, which implies
\begin{equation}
\hat{S}_M \stackrel{d}{=} \hat{S}_C.
\end{equation}

Furthermore, note that the variance of the empirical estimate \(\hat{F}_P(x)\) is solely a function of \(P\):
\begin{equation}
\operatorname{Var}[\hat{F}_P(x)] = \frac{F(x)(1-F(x))}{P}.
\end{equation}
Thus, when \(P\) changes (e.g., \(P=50\), \(100\), or \(500\)), the change in variance—and hence the fluctuation in the score—is proportional to \(\frac{1}{P}\) and is independent of the corpus. In other words,
\begin{equation}
\Delta \operatorname{Var} \propto \frac{1}{P},
\end{equation}
which holds for both the medical and the computer science datasets.

Therefore, we conclude that:
\begin{equation}
\boxed{\hat{S}_M \stackrel{d}{=} \hat{S}_C \quad \text{and} \quad \Delta \hat{S} \propto \frac{1}{P}.}
\end{equation}

This establishes that the scoring distributions for the test set are identical across corporas, and the effect of changing \(P\) on the score variation is equivalent for all datasets.

\subsubsection{Experimental Evidence}
\begin{figure}[H]
  \centering
  \includegraphics[width=\textwidth]{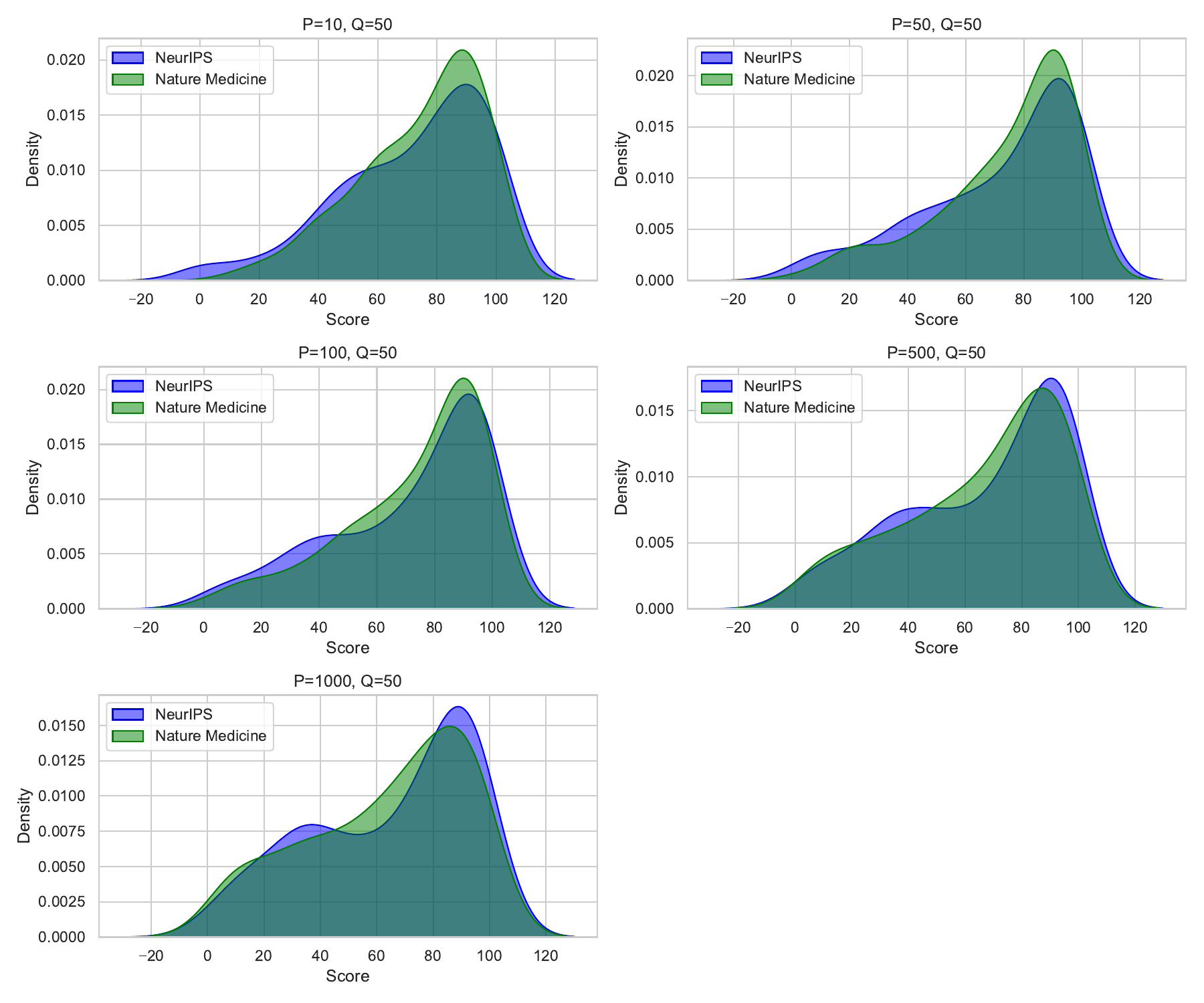}
  \caption{\textbf{Score Distribution of RND algorithm with different \(P\) value}.}
  \label{fig: Score distribution with dif P}
\end{figure}

\section{Test Set}
\subsection{NeurIPS Test Set}
\renewcommand{\arraystretch}{1.2}
\begin{longtable}{p{7cm} p{7cm}}
\caption{Titles of Novel (Positive) and Non-novel (Negative) Papers in NeurIPS Test Set}
\label{tab:NIPS_Dataset} \\

\multicolumn{2}{c}{\textbf{NeurIPS Test Set}} \\
\toprule
\multicolumn{1}{c}{\textbf{Positive}} & \multicolumn{1}{c}{\textbf{Negative}} \\
\midrule
\endfirsthead

\multicolumn{2}{c}{\textbf{NeurIPS Test Set}} \\
\toprule
\multicolumn{1}{c}{\textbf{Positive}} & \multicolumn{1}{c}{\textbf{Negative}} \\
\midrule
\endhead

\midrule
\multicolumn{2}{r}{{Continued on next page}} \\
\midrule
\endfoot

\endlastfoot
1. Learning to grok: Emergence of in-context learning and skill composition in modular arithmetic tasks & 1. Attention is All you Need \\

2. Nonlocal Attention Operator: Materializing Hidden Knowledge Towards Interpretable Physics Discovery & 2. PyTorch: An Imperative Style, High-Performance Deep Learning Library \\

3. Emergence of Hidden Capabilities: Exploring Learning Dynamics in Concept Space & 3. Language Models are Few-Shot Learners \\

4. Continual learning with the neural tangent ensemble & 4. A Unified Approach to Interpreting Model Predictions \\

5. Neglected Hessian component explains mysteries in sharpness regularization & 5. Inductive Representation Learning on Large Graphs \\

6. Generalization Analysis for Label-Specific Representation Learning & 6. Denoising Diffusion Probabilistic Models \\

7. The Power of Resets in Online Reinforcement Learning & 7. GANs Trained by a Two Time-Scale Update Rule Converge to a Local Nash Equilibrium \\

8. Paths to Equilibrium in Games & 8. PointNet++: Deep Hierarchical Feature Learning on Point Sets in a Metric Space \\

9. Double-Ended Synthesis Planning with Goal-Constrained Bidirectional Search & 9. LightGBM: A Highly Efficient Gradient Boosting Decision Tree \\

10. Time-Reversal Provides Unsupervised Feedback to LLMs & 10. Improved Training of Wasserstein GANs \\

11. Compositional Generalization Across Distributional Shifts with Sparse Tree Operations & 11. Improved Techniques for Training GANs \\

12. Stable Minima Cannot Overfit in Univariate ReLU Networks: Generalization by Large Step Sizes & 12. XLNet: Generalized Autoregressive Pretraining for Language Understanding \\

13. Rule Extrapolation in Language Modeling: A Study of Compositional Generalization on OOD Prompts & 13. Prototypical Networks for Few-shot Learning \\

14. A generalized neural tangent kernel for surrogate gradient learning & 14. Convolutional LSTM Network: A Machine Learning Approach for Precipitation Nowcasting \\

15. GREATS: Online Selection of High-Quality Data for LLM Training in Every Iteration & 15. Convolutional Neural Networks on Graphs with Fast Localized Spectral Filtering \\

16. Non-Asymptotic Uncertainty Quantification in High-Dimensional Learning & 16. Spatial Transformer Networks \\

17. Boosting Vision-Language Models with Transduction & 17. Matching Networks for One Shot Learning \\

18. Input-to-State Stable Coupled Oscillator Networks for Closed-form Model-based Control in Latent Space & 18. Learning both Weights and Connections for Efficient Neural Network \\

19. Assouad, Fano, and Le Cam with Interaction: A Unifying Lower Bound Framework and Characterization for Bandit Learnability & 19. Bootstrap Your Own Latent: A New Approach to Self-Supervised Learning \\

20. Exploring Jacobian Inexactness in Second-Order Methods for Variational Inequalities: Lower Bounds, Optimal Algorithms and Quasi-Newton Approximations & 20. Character-level Convolutional Networks for Text Classification \\

21. Who's asking? User personas and the mechanics of latent misalignment & 21. R-FCN: Object Detection via Region-based Fully Convolutional Networks \\

22. Self-Consuming Generative Models with Curated Data Provably Optimize Human Preferences & 22. Simple and Scalable Predictive Uncertainty Estimation using Deep Ensembles \\

23. Selective Generation for Controllable Language Models & 23. wav2vec 2.0: A Framework for Self-Supervised Learning of Speech Representations \\

24. Constrained Adaptive Attack: Effective Adversarial Attack Against Deep Neural Networks for Tabular Data & 24. Neural Ordinary Differential Equations \\

25. Learning Generalized Linear Programming Value Functions & 25. Dynamic Routing Between Capsules \\

26. Optimizing Automatic Differentiation with Deep Reinforcement Learning & 26. What Uncertainties Do We Need in Bayesian Deep Learning for Computer Vision? \\

27. Overcoming Common Flaws in the Evaluation of Selective Classification Systems & 27. Retrieval-Augmented Generation for Knowledge-Intensive NLP Tasks \\

28. Revisiting K-mer Profile for Effective and Scalable Genome Representation Learning & 28. Neural Discrete Representation Learning \\

29. Trading Place for Space: Increasing Location Resolution Reduces Contextual Capacity in Hippocampal Codes & 29. InfoGAN: Interpretable Representation Learning by Information Maximizing Generative Adversarial Nets \\

30. Reproducibility of predictive networks for mouse visual cortex & 30. Multi-Agent Actor-Critic for Mixed Cooperative-Competitive Environments \\

31. Nonlinear dynamics of localization in neural receptive fields & 31. Supervised Contrastive Learning \\

32. Learning Noisy Halfspaces with a Margin: Massart is No Harder than Random & 32. Equality of Opportunity in Supervised Learning \\

33. Cracking the Code of Juxtaposition: Can AI Models Understand the Humorous Contradictions & 33. Unsupervised Learning of Visual Features by Contrasting Cluster Assignments \\

34. Evaluating the World Model Implicit in a Generative Model & 34. Mean teachers are better role models: Weight-averaged consistency targets improve semi-supervised deep learning results \\

35. TrackIME: Enhanced Video Point Tracking via Instance Motion Estimation & 35. Teaching Machines to Read and Comprehend \\

36. DiffLight: A Partial Rewards Conditioned Diffusion Model for Traffic Signal Control with Missing Data & 36. ViLBERT: Pretraining Task-Agnostic Visiolinguistic Representations for Vision-and-Language Tasks \\

37. Mean-Field Langevin Dynamics for Signed Measures via a Bilevel Approach & 37. Convolutional Networks on Graphs for Learning Molecular Fingerprints \\

38. Stabilized Proximal-Point Methods for Federated Optimization & 38. Generative Modeling by Estimating Gradients of the Data Distribution \\

39. Reparameterization invariance in approximate Bayesian inference & 39. FixMatch: Simplifying Semi-Supervised Learning with Consistency and Confidence \\

40. Disentangling the Roles of Distinct Cell Classes with Cell-Type Dynamical Systems & 40. Learning Structured Output Representation using Deep Conditional Generative Models \\

41. Linear Regression using Heterogeneous Data Batches & 41. Glow: Generative Flow with Invertible 1x1 Convolutions \\

42. A Near-optimal Algorithm for Learning Margin Halfspaces with Massart Noise & 42. CatBoost: unbiased boosting with categorical features \\

43. Neural Krylov Iteration for Accelerating Linear System Solving & 43. Man is to Computer Programmer as Woman is to Homemaker? Debiasing Word Embeddings \\

44. Human Expertise in Algorithmic Prediction & 44. Neural Tangent Kernel: Convergence and Generalization in Neural Networks \\

45. Analysing Multi-Task Regression via Random Matrix Theory with Application to Time Series Forecasting & 45. Generative Adversarial Imitation Learning \\

46. No-regret Learning in Harmonic Games: Extrapolation in the Face of Conflicting Interests & 46. Pointer Networks \\

47. Learning diffusion at lightspeed & 47. BinaryConnect: Training Deep Neural Networks with binary weights during propagations \\

48. Voila-A: Aligning Vision-Language Models with User's Gaze Attention & 48. MixMatch: A Holistic Approach to Semi-Supervised Learning \\

49. Barely Random Algorithms and Collective Metrical Task Systems & 49. Unsupervised Image-to-Image Translation Networks \\

50. Goal Reduction with Loop-Removal Accelerates RL and Models Human Brain Activity in Goal-Directed Learning & 50. Deep Reinforcement Learning from Human Preferences \\

51. BricksRL: A Platform for Democratizing Robotics and  Reinforcement Learning Research and Education with LEGO & 51. Cross-lingual Language Model Pretraining \\

52. Breaking Long-Tailed Learning Bottlenecks: A Controllable Paradigm with Hypernetwork-Generated Diverse Experts & 52. Attention-Based Models for Speech Recognition \\

53. Kermut: Composite kernel regression for protein variant effects & 53. End-To-End Memory Networks \\

54. Automatically Learning Hybrid Digital Twins of Dynamical Systems & 54. Gradient Episodic Memory for Continual Learning \\

55. On the Identifiability of Poisson Branching Structural Causal Model Using Probability Generating Function & 55. Open Graph Benchmark: Datasets for Machine Learning on Graphs \\

56. Weisfeiler and Leman Go Loopy: A New Hierarchy for Graph Representational Learning & 56. Conditional Image Generation with PixelCNN Decoders \\

57. Unlocking the Capabilities of Thought: A Reasoning Boundary Framework to Quantify and Optimize Chain-of-Thought & 57. Generalized Cross Entropy Loss for Training Deep Neural Networks with Noisy Labels \\

58. Reinforcement Learning Under Latent Dynamics: Toward Statistical and Algorithmic Modularity & 58. Skip-Thought Vectors \\

59. Generalization Error Bounds for Two-stage Recommender Systems with Tree Structure & 59. Self-Normalizing Neural Networks \\

60. Can Transformers Smell Like Humans? & 60. PointCNN: Convolution On X-Transformed Points \\

61. Geodesic Optimization for Predictive Shift Adaptation on EEG data & 61. Learning Structured Sparsity in Deep Neural Networks \\

62. Second-order forward-mode optimization of recurrent neural networks for neuroscience & 62. Implicit Neural Representations with Periodic Activation Functions \\

63. Discrete Flow Matching & 63. Deep Generative Image Models using a Laplacian Pyramid of Adversarial Networks \\

64. Motion Forecasting in Continuous Driving & 64. Hindsight Experience Replay \\

65. Moving Off-the-Grid: Scene-Grounded Video Representations & 65. Unsupervised Data Augmentation for Consistency Training \\

66. Aligner: Efficient Alignment by Learning to Correct & 66. Conditional Adversarial Domain Adaptation \\

67. Questioning the Survey Responses of Large Language Models & 67. SuperGLUE: A Stickier Benchmark for General-Purpose Language Understanding Systems \\

68. Saliency-driven Experience Replay for Continual Learning & 68. Big Self-Supervised Models are Strong Semi-Supervised Learners \\

69. Adversarial Environment Design via Regret-Guided Diffusion Models & 69. Fourier Features Let Networks Learn High Frequency Functions in Low Dimensional Domains \\

70. Localized Zeroth-Order Prompt Optimization & 70. Improved Deep Metric Learning with Multi-class N-pair Loss Objective \\

71. Molecule Design by Latent Prompt Transformer & 71. Hierarchical Graph Representation Learning with Differentiable Pooling \\

72. Can Learned Optimization Make Reinforcement Learning Less Difficult? & 72. Scheduled Sampling for Sequence Prediction with Recurrent Neural Networks \\

73. Reverse Transition Kernel: A Flexible Framework to Accelerate Diffusion Inference & 73. Learning to learn by gradient descent by gradient descent \\

74. Any2Graph: Deep End-To-End Supervised Graph Prediction With An Optimal Transport Loss & 74. Co-teaching: Robust training of deep neural networks with extremely noisy labels \\

75. MInference 1.0: Accelerating Pre-filling for Long-Context LLMs via Dynamic Sparse Attention & 75. Continual Learning with Deep Generative Replay \\

76. Ensemble Learning for Heterogeneous Large Language Models with Deep Parallel Collaboration & 76. Learning a Probabilistic Latent Space of Object Shapes via 3D Generative-Adversarial Modeling \\

77. Humanoid Locomotion as Next Token Prediction & 77. Big Bird: Transformers for Longer Sequences \\

78. NeoRL: Efficient Exploration for Nonepisodic RL & 78. Weight Normalization: A Simple Reparameterization to Accelerate Training of Deep Neural Networks \\

79. Toxicity Detection for Free & 79. A Simple Unified Framework for Detecting Out-of-Distribution Samples and Adversarial Attacks \\

80. Semi-supervised Multi-label Learning with Balanced Binary Angular Margin Loss & 80. Sanity Checks for Saliency Maps \\

& 81. When Does Label Smoothing Help? \\

& 82. Graph Contrastive Learning with Augmentations \\

& 83. Link Prediction Based on Graph Neural Networks \\

& 84. Improved Variational Inference with Inverse Autoregressive Flow \\

& 85. Training Generative Adversarial Networks with Limited Data \\

& 86. Visualizing the Loss Landscape of Neural Nets \\

& 87. Adversarial Examples Are Not Bugs, They Are Features \\

& 88. HiFi-GAN: Generative Adversarial Networks for Efficient and High Fidelity Speech Synthesis \\

& 89. Understanding the Effective Receptive Field in Deep Convolutional Neural Networks \\

& 90. Learning to summarize from human feedback \\

& 91. Federated Multi-Task Learning \\

& 92. Training Very Deep Networks \\

& 93. Efficient and Robust Automated Machine Learning \\

& 94. A Theoretically Grounded Application of Dropout in Recurrent Neural Networks \\

& 95. Generating Diverse High-Fidelity Images with VQ-VAE-2 \\

& 96. f-GAN: Training Generative Neural Samplers using Variational Divergence Minimization \\

& 97. Coupled Generative Adversarial Networks \\

& 98. Conservative Q-Learning for Offline Reinforcement Learning \\

& 99. A simple neural network module for relational reasoning \\

\bottomrule
\end{longtable}

\subsection{Nature Medicine Test Set} 
\begin{longtable}{p{7cm} p{7cm}}
\label{tab:Nature_Medicine_Dataset} \\

\toprule
\multicolumn{2}{c}{\textbf{Nature Medicine Test Set}} \\
\toprule
\multicolumn{1}{c}{\textbf{Positive}} & \multicolumn{1}{c}{\textbf{Negative}} \\
\midrule
\endfirsthead

\multicolumn{2}{c}{\textbf{Nature Medicine Test Set}} \\
\toprule
\multicolumn{1}{c}{\textbf{Positive}} & \multicolumn{1}{c}{\textbf{Negative}} \\
\midrule
\endhead

\midrule
\multicolumn{2}{r}{{Continued on next page}} \\
\midrule
\endfoot

\endlastfoot
1. Sleep patterns and risk of chronic disease as measured by long-term monitoring with commercial wearable devices in the All of Us Research Program. & 1. Microenvironmental regulation of tumor progression and metastasis \\

2. Botensilimab plus balstilimab in relapsed/refractory microsatellite stable metastatic colorectal cancer: a phase 1 trial. & 2. Temporal dynamics in viral shedding and transmissibility of COVID-19 \\

3. Lipidome changes due to improved dietary fat quality inform cardiometabolic risk reduction and precision nutrition. & 3. The Consensus Molecular Subtypes of Colorectal Cancer \\

4. Fratricide-resistant CD7-CAR T cells in T-ALL. & 4. High-performance medicine: the convergence of human and artificial intelligence \\

5. International multicenter validation of AI-driven ultrasound detection of ovarian cancer. & 5. Understanding the tumor immune microenvironment (TIME) for effective therapy \\

6. Donor-derived GD2-specific CAR T cells in relapsed or refractory neuroblastoma. & 6. Intestinal microbiota metabolism of L-carnitine, a nutrient in red meat, promotes atherosclerosis \\

7. Single-nucleus chromatin accessibility and transcriptomic map of breast tissues of women of diverse genetic ancestry. & 7. Post-acute COVID-19 syndrome \\

8. Unidirectional association of clonal hematopoiesis with atherosclerosis development. & 8. Melanoma exosomes educate bone marrow progenitor cells toward a pro-metastatic phenotype through MET \\

9. Echocardiographic screening for heart failure and optimization of the care pathway for individuals with pacemakers: a randomized controlled trial. & 9. Signatures of T cell dysfunction and exclusion predict cancer immunotherapy response \\

10. Population-based, first-tier genomic newborn screening in the maternity ward. & 10. Neutralizing antibody levels are highly predictive of immune protection from symptomatic SARS-CoV-2 infection \\

11. Allogeneic CD5-specific CAR-T therapy for relapsed/refractory T-ALL: a phase 1 trial. & 11. Ischemia and reperfusion—from mechanism to translation \\

12. Transplantation of a genetically modified porcine heart into a live human. & 12. Mechanisms of fibrosis: therapeutic translation for fibrotic disease \\

13. A multi-modal single-cell and spatial expression map of metastatic breast cancer biopsies across clinicopathological features. & 13. Metabolite profiles and the risk of developing diabetes \\

14. ctDNA-based molecular residual disease and survival in resectable colorectal cancer. & 14. Mechanisms of NAFLD development and therapeutic strategies \\

15. Antifungal heteroresistance causes prophylaxis failure and facilitates breakthrough Candida parapsilosis infections. & 15. Inflammasomes: mechanism of action, role in disease, and therapeutics \\

16. Subcutaneous weekly semaglutide with automated insulin delivery in type 1 diabetes: a double-blind, randomized, crossover trial. & 16. Chronic inflammation in the etiology of disease across the life span \\

17. Combined endurance and resistance exercise training in heart failure with preserved ejection fraction: a randomized controlled trial. & 17. Mutational Landscape of Metastatic Cancer Revealed from Prospective Clinical Sequencing of 10,000 Patients \\

18. Multi-omic profiling a defined bacterial consortium for treatment of recurrent Clostridioides difficile infection. & 18. Antibody responses to SARS-CoV-2 in patients with COVID-19 \\

19. An organotypic atlas of human vascular cells. & 19. ABT-199, a potent and selective BCL-2 inhibitor, achieves antitumor activity while sparing platelets \\

20. Lipid profiling identifies modifiable signatures of cardiometabolic risk in children and adolescents with obesity. & 20. Clinical and immunological assessment of asymptomatic SARS-CoV-2 infections \\

21. Ferric carboxymaltose for anemia in late pregnancy: a randomized controlled trial. & 21. Extrapulmonary manifestations of COVID-19 \\

22. Effects of conditional cash transfers on tuberculosis incidence and mortality according to race, ethnicity and socioeconomic factors in the 100 Million Brazilian Cohort. & 22. A guide to deep learning in healthcare \\

23. Phenome-wide associations of sleep characteristics in the Human Phenotype Project. & 23. A global survey of potential acceptance of a COVID-19 vaccine \\

24. Proteomic signatures improve risk prediction for common and rare diseases. & 24. The emerging role of lncRNAs in cancer \\

25. Remotely delivered weight management for people with long COVID and overweight: the randomized wait-list-controlled ReDIRECT trial. & 25. SARS-CoV-2 Entry Genes Are Most Highly Expressed in Nasal Goblet and Ciliated Cells within Human Airways \\

26. Sustained effect of prasinezumab on Parkinson's disease motor progression in the open-label extension of the PASADENA trial. & 26. Gut microbiota metabolism of dietary fiber influences allergic airway disease and hematopoiesis \\

27. Collaboration between clinicians and vision-language models in radiology report generation. & 27. The immunology of stroke: from mechanisms to translation \\

28. Oral obeldesivir provides postexposure protection against Marburg virus in nonhuman primates. & 28. Asthma phenotypes: the evolution from clinical to molecular approaches \\

29. Digital consults in heart failure care: a randomized controlled trial. & 29. Single-cell landscape of bronchoalveolar immune cells in patients with COVID-19 \\

30. Increased frequency of repeat expansion mutations across different populations. & 30. A small-molecule inhibitor of the NLRP3 inflammasome for the treatment of inflammatory diseases \\

31. Autogene cevumeran with or without atezolizumab in advanced solid tumors: a phase 1 trial. & 31. Respiratory virus shedding in exhaled breath and efficacy of face masks \\

32. A high-performance brain-computer interface for finger decoding and quadcopter game control in an individual with paralysis. & 32. Cancer stem cells revisited \\

33. Mapping the effectiveness and risks of GLP-1 receptor agonists. & 33. Atherosclerosis: current pathogenesis and therapeutic options \\

34. Evaluating generalizability of oncology trial results to real-world patients using machine learning-based trial emulations. & 34. Brown and beige fat: development, function and therapeutic potential \\

35. AI-based differential diagnosis of dementia etiologies on multimodal data. & 35. Cardiologist-level arrhythmia detection and classification in ambulatory electrocardiograms using a deep neural network \\

36. Molecular classification to refine surgical and radiotherapeutic decision-making in meningioma. & 36. An inflammatory cytokine signature predicts COVID-19 severity and survival \\

37. A framework for sharing of clinical and genetic data for precision medicine applications. & 37. Classification and mutation prediction from non–small cell lung cancer histopathology images using deep learning \\

38. A generalist medical language model for disease diagnosis assistance. & 38. CSF-1R inhibition alters macrophage polarization and blocks glioma progression \\

39. Subclassification of obesity for precision prediction of cardiometabolic diseases. & 39. Clinically applicable deep learning for diagnosis and referral in retinal disease \\

40. Somatic CAG repeat expansion in blood associates with biomarkers of neurodegeneration in Huntington's disease decades before clinical motor diagnosis. & 40. Development, maintenance and disruption of the blood-brain barrier \\

41. Genomic reanalysis of a pan-European rare-disease resource yields new diagnoses. & 41. Adipocytes promote ovarian cancer metastasis and provide energy for rapid tumor growth \\

42. In vivo base editing extends lifespan of a humanized mouse model of prion disease. & 42. An ultrasensitive method for quantitating circulating tumor DNA with broad patient coverage \\

43. Self-improving generative foundation model for synthetic medical image generation and clinical applications. & 43. The cancer stem cell: premises, promises and challenges \\

44. Data-driven cluster analysis identifies distinct types of metabolic dysfunction-associated steatotic liver disease. & 44. Attributes and predictors of long COVID \\

45. The economic value of reducing avoidable mortality. & 45. WNT signaling in bone homeostasis and disease: from human mutations to treatments \\

46. Genetic basis of early onset and progression of type 2 diabetes in South Asians. & 46. The role of autophagy in neurodegenerative disease \\

47. Posthospitalization COVID-19 cognitive deficits at 1 year are global and associated with elevated brain injury markers and gray matter volume reduction. & 47. Clinical-grade computational pathology using weakly supervised deep learning on whole slide images \\

48. Safety and reactogenicity of a controlled human infection model of sand fly-transmitted cutaneous leishmaniasis. & 48. A human memory T-cell subset with stem cell-like properties \\

49. Cabozantinib and nivolumab with or without live bacterial supplementation in metastatic renal cell carcinoma: a randomized phase 1 trial. & 49. Current understanding of the human microbiome \\

50. Seven-year performance of a clinical metagenomic next-generation sequencing test for diagnosis of central nervous system infections. & 50. Molecular analysis of gastric cancer identifies subtypes associated with distinct clinical outcomes \\

51. Brain aging patterns in a large and diverse cohort of 49,482 individuals. & 51. Cellular senescence in aging and age-related disease: from mechanisms to therapy \\

52. Large floods drive changes in cause-specific mortality in the United States. & 52. PPAR$\gamma$ signaling and metabolism: the good, the bad and the future \\

53. Cytokine-mediated CAR T therapy resistance in AML. & 53. Resting-state connectivity biomarkers define neurophysiological subtypes of depression \\

54. Prediction of brain metastasis development with DNA methylation signatures. & 54. Age-dependent effects in the transmission and control of COVID-19 epidemics \\

55. Personalized, autologous neoantigen-specific T cell therapy in metastatic melanoma: a phase 1 trial. & 55. Tumor angiogenesis: molecular pathways and therapeutic targets \\

56. A generalist vision-language foundation model for diverse biomedical tasks. & 56. Evidence for osteocyte regulation of bone homeostasis through RANKL expression \\

57. DNA liquid biopsy-based prediction of cancer-associated venous thromboembolism. & 57. Modelling the COVID-19 epidemic and implementation of population-wide interventions in Italy \\

58. Semaglutide in patients with overweight or obesity and chronic kidney disease without diabetes: a randomized double-blind placebo-controlled clinical trial. & 58. Identification of the molecular basis of doxorubicin-induced cardiotoxicity \\

59. Intracerebroventricular B7-H3-targeting CAR T cells for diffuse intrinsic pontine glioma: a phase 1 trial. & 59. New from NPG: Genome-wide association study identifies five new schizophrenia loci \\

60. AI-based selection of individuals for supplemental MRI in population-based breast cancer screening: the randomized ScreenTrustMRI trial. & 60. Senolytics Improve Physical Function and Increase Lifespan in Old Age \\

61. A toolbox for surfacing health equity harms and biases in large language models. & 61. Subtypes of Pancreatic Ductal Adenocarcinoma and Their Differing Responses to Therapy \\

62. Partitioned polygenic risk scores identify distinct types of metabolic dysfunction-associated steatotic liver disease. & 62. A purified membrane protein from Akkermansia muciniphila or the pasteurized bacterium improves metabolism in obese and diabetic mice \\

63. Multi-omics-based mapping of decidualization resistance in patients with a history of severe preeclampsia. & 63. The NALP3/NLRP3 Inflammasome Instigates Obesity-Induced Autoinflammation and Insulin Resistance \\

64. Electronic nudges for sustained influenza vaccination uptake in older adults: the nationwide randomized NUDGE-FLU-2 trial. & 64. IgE and mast cells in allergic disease \\

65. A time-stratified, case-crossover study of heat exposure and perinatal mortality from 16 hospitals in sub-Saharan Africa. & 65. Brown adipose tissue activity controls triglyceride clearance \\

66. SARS-CoV-2 correlates of protection from infection against variants of concern. & 66. Intraoperative tumor-specific fluorescence imaging in ovarian cancer by folate receptor-$\alpha$ targeting: first in-human results \\

& 67. The cellular and signaling networks linking the immune system and metabolism in disease \\

& 68. Supplementation with Akkermansia muciniphila in overweight and obese human volunteers: a proof-of-concept exploratory study \\

& 69. Why don't we get more cancer? A proposed role of the microenvironment in restraining cancer progression \\

& 70. Clearance of senescent cells by ABT263 rejuvenates aged hematopoietic stem cells in mice \\

& 71. 4-1BB Costimulation Ameliorates T Cell Exhaustion Induced by Tonic Signaling of Chimeric Antigen Receptors \\

& 72. Characteristics of pediatric SARS-CoV-2 infection and potential evidence for persistent fecal viral shedding \\

& 73. Molecular subtypes of diffuse large B cell lymphoma are associated with distinct pathogenic mechanisms and outcomes \\

& 74. The oral and gut microbiomes are perturbed in rheumatoid arthritis and partly normalized after treatment \\

& 75. End-to-end lung cancer screening with three-dimensional deep learning on low-dose chest computed tomography \\

& 76. The practical implementation of artificial intelligence technologies in medicine \\

& 77. Estimating clinical severity of COVID-19 from the transmission dynamics in Wuhan, China \\

& 78. Microglia emerge as central players in brain disease \\

& 79. Organ reengineering through development of a transplantable recellularized liver graft using decellularized liver matrix \\

& 80. RET, ROS1 and ALK fusions in lung cancer \\

& 81. Divergent clonal evolution of castration resistant neuroendocrine prostate cancer \\

& 82. Long-term cardiovascular outcomes of COVID-19 \\

& 83. Ketone body $\beta$-hydroxybutyrate blocks the NLRP3 inflammasome-mediated inflammatory disease \\

& 84. Large language models in medicine \\

& 85. In vivo photodynamic therapy using upconversion nanoparticles as remote-controlled nanotransducers \\

& 86. Mitochondrial transfer from bone-marrow–derived stromal cells to pulmonary alveoli protects against acute lung injury \\

& 87. A single-cell atlas of the peripheral immune response in patients with severe COVID-19 \\

& 88. Determinants of response and resistance to CD19 chimeric antigen receptor (CAR) T cell therapy of chronic lymphocytic leukemia \\

& 89. Cancer epigenetics reaches mainstream oncology \\

& 90. Real-time tracking of self-reported symptoms to predict potential COVID-19 \\

& 91. Metformin alters the gut microbiome of individuals with treatment-naive type 2 diabetes, contributing to the therapeutic effects of the drug \\

& 92. Synaptic plasticity and depression: new insights from stress and rapid-acting antidepressants \\

& 93. Matrix-embedded cells control osteoclast formation \\

& 94. Targeting EZH2 in cancer \\

& 95. Comprehensive molecular characterization of clinical responses to PD-1 inhibition in metastatic gastric cancer \\

& 96. Identification of miR-34a as a potent inhibitor of prostate cancer progenitor cells and metastasis by directly repressing CD44 \\

& 97. Phenotype molding of stromal cells in the lung tumor microenvironment \\

& 98. Key roles of adjuvants in modern vaccines \\

& 99. AI in health and medicine \\
\bottomrule
\caption{Titles of Novel (Positive) and Non-novel (Negative) Papers in Nature Medicine Test Set} \\

\end{longtable}
\renewcommand{\arraystretch}{1.0}

\section{Prompt}
\subsection{Prompt for LLM with NeurIPS 2024 Review Guideline}
\begin{longtable}{p{14cm}}
\label{tab:LLM_with_guideline_prompt} \\

\toprule
\textbf{Prompt} \\
\midrule
\textbf{Task description}: You are a researcher who is reviewing a paper that was submitted to a computer science venue. Be critical and cautious
in your decision. If a paper is bad or you are unsure, give it bad scores and reject it. Below is a description of the
questions you will be asked on the review form for each paper and some guidelines on what to consider when
answering these questions.\\
\textbf{Reviewer guidelines}:
1. Summary: Briefly summarize the paper and its contributions. This is not the place to critique the paper; the authors should generally agree with a well-written summary.\\
2. Strengths and Weaknesses: Please provide a thorough assessment of the strengths and weaknesses of the paper,
touching on each of the following dimensions:\\
     - Originality: Are the tasks or methods new? Is the work a novel combination of well-known techniques? (This can be valuable!) Is it clear how this work differs from previous contributions?\\
     - Quality: Is the submission technically sound? Are claims well-supported (e.g., by theoretical analysis or experimental results)? Are the methods used appropriately? Is this a complete piece of work or a work in progress? Are the authors careful and honest about evaluating both the strengths and weaknesses of their work?\\
     - Clarity: Is the submission clearly written? Is it well organized? (If not, please make constructive suggestions for improving its clarity.) Does it adequately inform the reader? (Note that a superbly written paper provides enough information for an expert reader to reproduce its results.)\\
     - Significance: Are the results important? Are others (researchers or practitioners) likely to use the ideas or build on them? Does the submission address a difficult task in a better way than previous work? Does it advance the state of the art in a demonstrable way? Does it provide unique data, unique conclusions about existing data, or a unique theoretical
    or experimental approach?\\
3. Questions: Please list and carefully describe any questions and suggestions for the authors. Think of the things where a response from the author can change your opinion, clarify confusion, or address a limitation. This can be very important for a productive rebuttal and discussion phase with the authors.\\
4. Ethical concerns: If there are ethical issues with this paper, please flag the paper for an ethics review.\\
5. Overall: Please provide an "overall score" for this submission. \\
    Choices:\\
     - 10: Award quality: Technically flawless paper with groundbreaking impact on one or more areas, with
    exceptionally strong evaluation, reproducibility, and resources, and no unaddressed ethical considerations.\\
     - 9: Very Strong Accept: Technically flawless paper with groundbreaking impact on at least one area and excellent
    impact on multiple areas, with flawless evaluation, resources, and reproducibility, and no unaddressed ethical
    considerations.\\
     - 8: Strong Accept: Technically strong paper, with novel ideas, excellent impact on at least one area or high-toexcellent impact on multiple areas, with excellent evaluation, resources, and reproducibility, and no unaddressed ethical
    considerations.\\
     - 7: Accept: Technically solid paper, with high impact on at least one sub-area or moderate-to-high impact on more
    than one area, with good-to-excellent evaluation, resources, reproducibility, and no unaddressed ethical considerations.\\
     - 6: Weak Accept: Technically solid, moderate-to-high impact paper, with no major concerns with respect to
    evaluation, resources, reproducibility, and ethical considerations.\\
     - 5: Borderline accept: Technically solid paper where reasons to accept outweigh reasons to reject, e.g., limited
    evaluation. Please use sparingly.\\
     - 4: Borderline reject: Technically solid paper where reasons to reject, e.g., limited evaluation, outweigh reasons to
    accept, e.g., good evaluation. Please use sparingly.\\
     - 3: Reject: For instance, a paper with technical flaws, weak evaluation, inadequate reproducibility, and incompletely
    addressed ethical considerations.\\
     - 2: Strong Reject: For instance, a paper with major technical flaws, and/or poor evaluation, limited impact, poor
    reproducibility, and mostly unaddressed ethical considerations.\\
     - 1: Very Strong Reject: For instance, a paper with trivial results or unaddressed ethical considerations\\

\textbf{Provided paper}:

Here is the paper you are asked to review: \\
\{paper\}

\textbf{Output}:\\
Return a JSON object:\\
<JSON>\\
{template}\\
<JSON>\\
\bottomrule
\\
\caption{Prompt for LLM with NeurIPS 2024 Review Guideline}
\end{longtable}

\subsection{Prompt for Standardized Project Proposals}
\begin{longtable}{p{14cm}}
\label{tab:Standardized Project Proposals prompt} \\

\toprule
\textbf{Prompt} \\
\midrule
\textbf{Role}: You are a writing assistant specialized in editing academic writing.\\
\textbf{Task}: I will give you a student’s research idea and an idea template. Your task is to edit the student’s idea to follow the template’s format.\\
\textbf{Student idea}:\\
Title\\
\{title\}\\
Main Idea\\
\{paper\}\\ 
\textbf{Template}:\\
1. Title: A concise statement of the main research question to be used as the paper title.\\
2. Problem Statement: Clearly define the problem your research intends to address. Explain clearly why this problem is interesting and important.\\
3. Motivation: Explain why existing methods are not good enough to solve the problem, and explain the inspiration behind the new proposed method. You should also motivate why the proposed method would work better than existing baselines on the problem.\\
4. Proposed Method: Explain how the proposed method works, describe all the essential steps.\\
5. Step-by-Step Experiment Plan: Break down every single step of the experiments, make sure every step is executable. Cover all essential details such as the datasets, models, and metrics to be used. If the project involves prompting, give some example prompts for each step.\\
6. Test Case Examples: Give at least two concrete examples. The first example should show how the baseline method fails on the test case. If there are multiple baselines, give examples for all of them. The second example should show how the proposed method succeeds on the test case. For each test case, include the input (test example and the full prompt) and the expected output. You should also
provide an explanation for why the outputs from the proposed prompt are better. If the proposed
method has multiple steps, break them down into intermediate steps.\\
7. Fallback Plan: Propose some alternative plans for what should the students do if the proposed method doesn’t manage to satisfy the success criteria. For example, you can suggest additional analysis to help debug why the proposed method didn’t work, which could inform alternative new methods, or just turn the project into an analysis paper instead by offering some interesting ablation
and insights.\\
\textbf{Requirement}:\\
Make sure that you only edit the wording and formatting, including things like punctuation, capitalization,
linebreaks, and bullet points. Also make sure to edit any informal wording and phrasing to use vocabulary that
sounds like the template’s writing style. No other changes are allowed beyond these.\\
You should use tab as indentation and make sure to use appropriate nested indentation for sub-bullets. All bullets
should have a clear hierarchy so people can easily differentiate the sub-bullets. Only leave empty lines between sections and remove any extra line breaks. If many bullet points are clustered together in a paragraph, separate them clearly with indentation and appropriate bullet point markers. Change to a new line for each new bullet point.\\
For the fallback plan, do not list a bunch of bullet points. Instead, condense them into one coherent paragraph.
For line breaks, avoid Raw String Literals or Double Backslashes when using "\\n", and change them to spaces or
tabs.\\
For in-line citations, if the citation mentioned the author’s last name (like "(Si et al., 2023)" or "(An et al., 2024)"), you should keep them there; but if the citation is just a number (like "[1]" or "[3,4,5]"), you should just remove it and do some necessary rephrasing to make the sentence still sound coherent without the references.\\
Apart from minor rephrasing and changing formatting, do not change any content of the idea. You must preserve
the exact meaning of the original idea, do not change, remove, or add any other details. Do not drop any sections
(including test case examples). Do not rename any models, datasets, or methods. Do not drop clarification or
examples in brackets and do not drop any data source mentions (e.g., Chatbot Arena or Wildchat)! Note that
when indexing test case examples, each test case example could have multiple steps of inputs and outputs and you
shouldn’t give separate indices to them. Each test case example should be a whole set of input-output pairs for the
baseline(s) and proposed method.\\
For the proposed method section, avoid any big changes. If the section comes in as a coherent paragraph, you don’t
have to break it down into bullet points. If the section is already in bullet points, you should keep it that way. If the section is a mix of both, you should keep the bullet points and the coherent paragraph as they are.
Keep all the clarification and examples mentioned in all the sections and do not remove any of them (including those
in brackets).\\
For model selection, if any version of Claude is mentioned, change it to the latest version of Claude (Claude-3.5); if
any version of LLaMA is mentioned, change it to the latest version LLaMA-3. Do not make any other model changes.
Now directly generate the edited student idea to match the format of the template.\\
\bottomrule
\\
\caption{Prompt for Standardized Project Proposals} \\
\end{longtable}

\subsection{Prompt for LLM with Literature Search}
\begin{longtable}{p{14cm}}
\label{tab:LLM_with_paper_prompt} \\

\toprule
\textbf{Prompt} \\
\midrule
\textbf{Role}: You are an ambitious AI PhD student who is looking to publish a paper that will contribute significantly to the field.\\
\textbf{Task description}:\\
You have an idea and you want to check if it is novel or not. I.e., not overlapping significantly with existing literature or already well explored. Be a harsh critic for novelty, ensure there is a sufficient contribution in the idea for a new conference or workshop paper.\\
You will be given the titles and abstracts of the 10 papers most relevant to your idea. Decide a paper idea is novel if after sufficient searching, you have not found a paper that significantly overlaps with your idea. Decide a paper idea is not novel, if you have found a paper that significantly overlaps with your idea.\\
Set your decision to True if you think the idea is novel, set it to False if you think the idea is not novel.\\

\textbf{Your Idea}:\\
This is the idea you need to judge for novelty: \\
\{Idea\}\\

\textbf{Top 10 relevant papers}:\\
\{papers\}\\

\textbf{Output}:\\
Return only True or False, dont return any other words.\\
\bottomrule
\\
\caption{Prompt for LLM with Literature Search}
\end{longtable}

\newpage

\section{Result in Detail}
\subsection{AUROC of Different LLM-Based Model in Detail}
\begin{table}[H]
\centering
\renewcommand{\arraystretch}{1.2}
\label{tab:AUROC of different LLM model in detail}
\begin{tabular}{ccccccc}
\toprule
& & \multirow{2}{*}{\textbf{\makecell{Sonnet-3.7\\with\\guideline}}} & \multirow{2}{*}{\textbf{\makecell{Sonnet-3.7\\with\\tournament}}}& \multicolumn{3}{c}{\textbf{LLM + literature search}}\\
\cmidrule(lr){5-7}
                    & & & & \textbf{Sonnet-3.7} & \textbf{Deepseek-r1} & \textbf{Gpt-4o}\\
\midrule
\multirow{3}{*}{Round1} & NeurIPS & 0.544 & 0.497 & 0.818 & 0.746 & 0.56 \\
& Nature Medicine & NaN & 0.501 & 0.616 & 0.663 & 0.518 \\
& Mixed & NaN & 0.5 & 0.596 & 0.583 & 0.492 \\
\midrule
\multirow{3}{*}{Round2} & NeurIPS & 0.505 & 0.496 & 0.799 & 0.701 & 0.578 \\
& Nature Medicine & NaN & 0.51 & 0.624 & 0.707 & 0.551 \\
& Mixed & NaN & 0.497 & 0.603 & 0.661 & 0.535 \\
\midrule
\multirow{3}{*}{Round3} & NeurIPS & 0.59 & 0.496 & 0.823 & 0.682 & 0.564 \\
& Nature Medicine & NaN & 0.497 & 0.609 & 0.649 & 0.566 \\
& Mixed & NaN & 0.506 & 0.593 & 0.543 & 0.54 \\
\bottomrule
\end{tabular}
\caption{AUROC of Different LLM-Based Model in Detail}
\end{table}
\renewcommand{\arraystretch}{1}

\subsection{Comparison of AUROC of RND algorithm with different P values}

\begin{table}[H]
    \centering
    \begin{tabular}{cccccc}
        \toprule
        \multirow{2}{*}{\textbf{Test set}} & \multicolumn{5}{c}{\textbf{AUROC}} \\  
        \cmidrule(lr){2-6}
        & P=10,Q=50 & P=50,Q=50 & P=100,Q=50 & P=500,Q=50 & P=1000,Q=50 \\
        \midrule
        NeurIPS & 0.808	& 0.830	& 0.820	& 0.825	& 0.833 \\
        Nature Medicine & 0.686	& 0.754	& 0.765	& 0.760	& 0.765 \\
        Mixed & 0.762 & 0.805 & 0.795 & 0.786 &	0.787 \\
        \bottomrule
    \end{tabular}
    \caption{Comparison of AUROC of RND algorithm with different P values}
\end{table}

\subsection{Comparison of AUROC of RND algorithm with different Q values}

\begin{table}[H]
    \centering
    \begin{tabular}{cccccccc}
        \toprule
        \multirow{2}{*}{\textbf{Test set}} & \multicolumn{7}{c}{\textbf{AUROC}} \\  
        \cmidrule(lr){2-8}
        & P=100,Q=2 & P=100,Q=5 & P=100,Q=10 & P=100,Q=50 & P=100,Q=100 & P=100,Q=500 & P=100,Q=1000\\
        \midrule
        NeurIPS & 0.778 & 0.818 & 0.827 & 0.820 & 0.808 & 0.788 & 0.768\\
        Nature Medicine & 0.603 & 0.686 & 0.723 & 0.765 & 0.757 & 0.744 & 0.722\\
        Mixed & 0.628 & 0.745 & 0.774 & 0.795 & 0.782 & 0.771 & 0.758\\
        \bottomrule
    \end{tabular}
    \caption{Comparison of AUROC of RND algorithm with different Q values}
\end{table}

\section{Computing Resources}
To conduct our experiment, the necessary computing resources includes

1. Literature database with embedding vector: we used a server with 16 CPU cores, 128GB RAM, 2TB SSD disk(which actually used around 1TB) to deploy an ElasticSearch version 8 as literature search engine. The embedding of each literature is obtained using public available API services.

2. Algorithm development and experiment: a server with 16 CPU cores, 128GB RAM.

\end{document}